\documentclass[10pt,twocolumn]{article}

\usepackage{hhline}
\usepackage{makeidx}
\usepackage{graphicx}
\usepackage{amsmath,amssymb} 
\usepackage{microtype}
\usepackage{amssymb}
\usepackage{subfig}
\usepackage{rotating}
\usepackage{natbib}
\usepackage{hyperref}

\usepackage{algorithm}
\usepackage[noend]{algpseudocode}
\usepackage{etoolbox}

\begin{document}

\title{Inference, Learning and Attention Mechanisms
 that Exploit and
 Preserve Sparsity in CNNs
\author{Timo Hackel\footnote{Shared first 
		authorship} \and
        Mikhail Usvyatsov\footnotemark[1] \and 
        Silvano Galliani \and  Jan D. 
        Wegner \and 
        Konrad Schindler
}
}

\maketitle

\makeatletter{\renewcommand*{\@makefnmark}{}
	\footnotetext{Photogrammetry \& Remote Sensing, ETH Zurich, Switzerland.
		Correspondence to: Mikhail Usvyatsov 
		\href{mailto:mikhail.usvyatsov@geod.baug.ethz.ch}{mikhail.usvyatsov@geod.baug.ethz.ch}.}
	\makeatother}

\begin{abstract}
	Convolutional neural networks (CNNs) are a powerful tool for pattern
	recognition and computer vision, but they do not scale well to
	higher-dimensional inputs, because of the associated memory demands
	for storing and manipulating high-dimensional tensors.
	This work starts from the observation that higher-dimensional data,
	like for example 3D voxel volumes, are sparsely populated.
	CNNs naturally lend themselves to densely sampled data, and
	sophisticated, massively parallel implementations are available. On
	the contrary, existing frameworks by and large lack the ability to
	efficiently process sparse data. Here, we introduce a suite of
	tools that exploit sparsity in both the feature maps and the filter
	weights of a CNN, and thereby allow for significantly lower memory
	footprints and computation times than the conventional dense
	framework, when processing data with a high degree of sparsity. Our
	scheme provides \emph{(i)}~an efficient GPU implementation of a
	convolution layer based on direct, sparse convolution, as well as
	sparse implementations of the \emph{ReLU} and $max$-pooling
	layers; \emph{(ii)}~a filter step within the convolution layer, which
	we call \emph{attention}, that prevents fill-in, i.e., the tendency of
	convolution to rapidly decrease sparsity, and guarantees an upper
	bound on the computational resources; and \emph{(iii)}~an adaptation
	of back-propagation that makes it possible to combine our approach
	with standard learning frameworks, while still benefitting from
	sparsity in the data as well as the model.
\end{abstract}

\section{Introduction}
\label{sec:introduction}

Deep convolutional neural networks (CNNs) are nowadays the
state-of-the-art tool for a wide spectrum of computer vision
task~\citep{krizhevsky2012imagenet,long2015fully,ren2015faster}.
A main reason for their spectacular comeback, perhaps even the single
most important factor, is that today the necessary computer hardware
is available to train and apply deep neural networks with millions of
tunable parameters.
Graphics Processing Units (GPUs) have brought about an enormous
speed-up for massively parallel computation tasks. In particular, this
includes both the training (error back-propagation) and inference
(forward pass) steps of neural networks, where both the response maps
(feature maps) within the network and the network parameters (filter
weights) form regular grids that are conveniently stored and processed
as tensors.

\begin{figure}[t!]
	\centering
	\begin{tabular}{cc}
		\includegraphics[width=0.4\columnwidth, trim=300 50 300 150,
		clip=true, keepaspectratio]{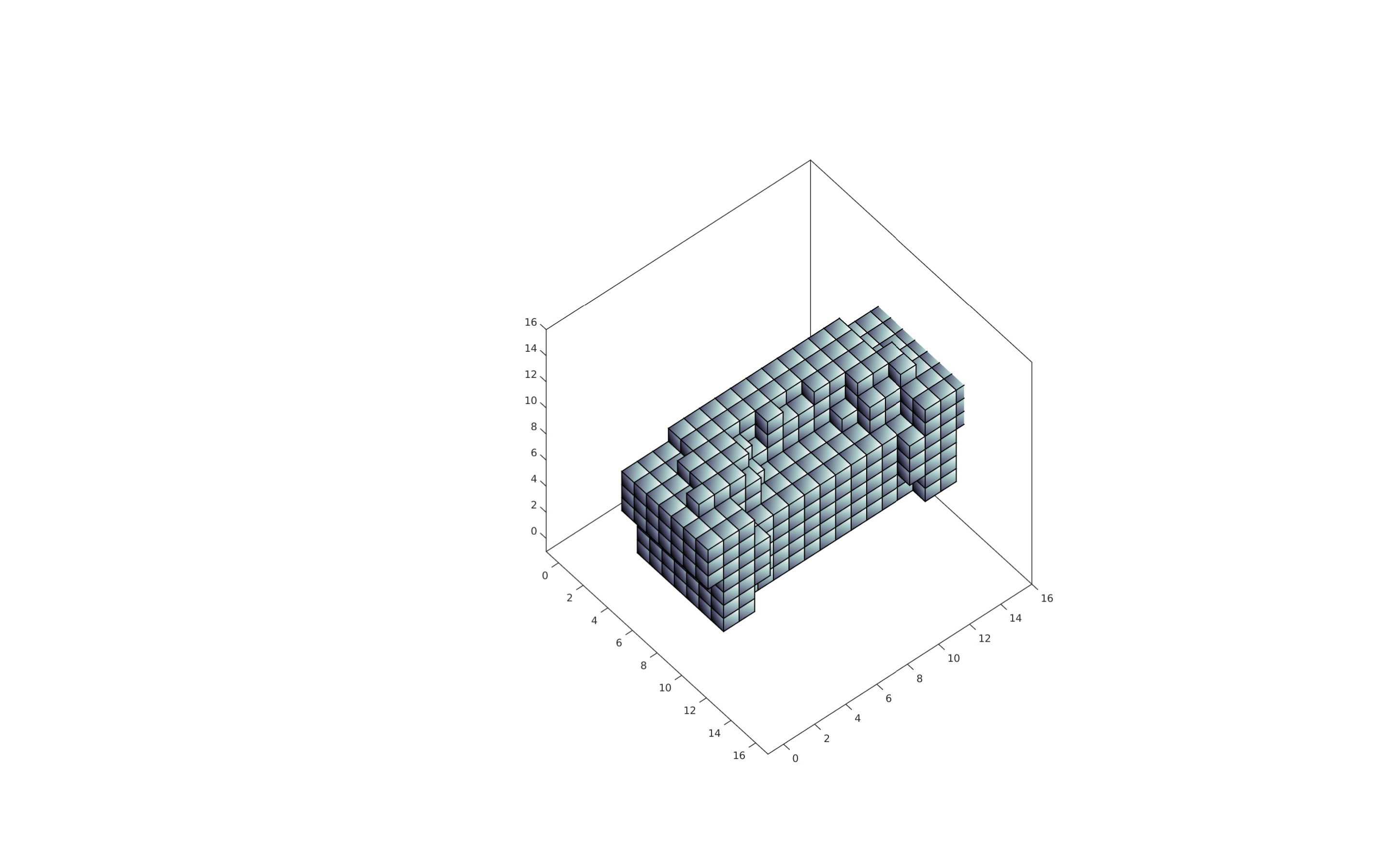} &
		\includegraphics[width=0.4\columnwidth, trim=300 30 300 150,
		clip=true, keepaspectratio]{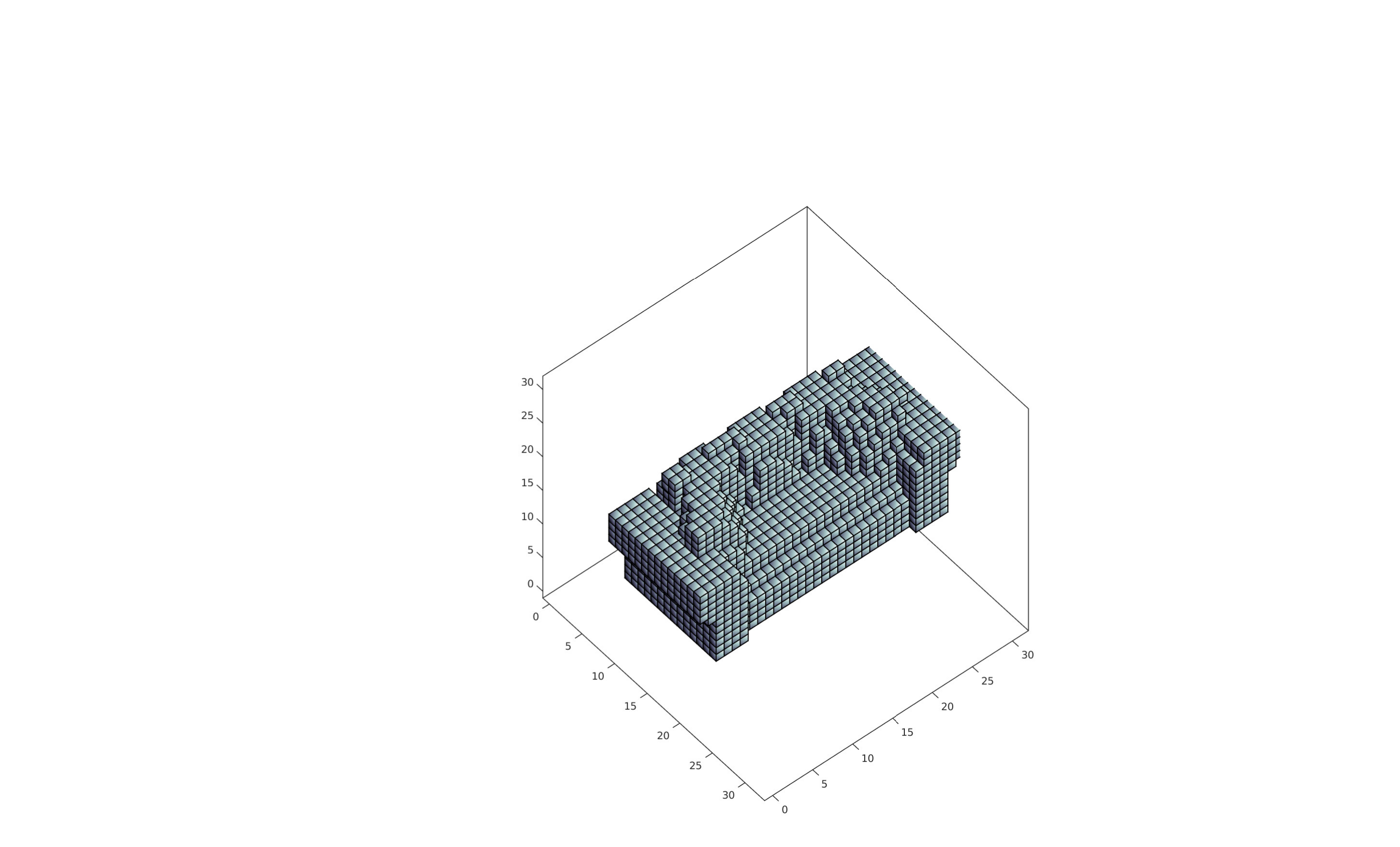} \\
		$\;\;\;\;r=16$ &
		$\;\;\;\;r=32$ \\
		$\;\;\;\;\overline\rho=15.5\%$ &
		$\;\;\;\;\overline\rho=9.1\%$ \\
		\includegraphics[width=0.4\columnwidth, trim=300 80 300 120,
		clip=true, keepaspectratio]{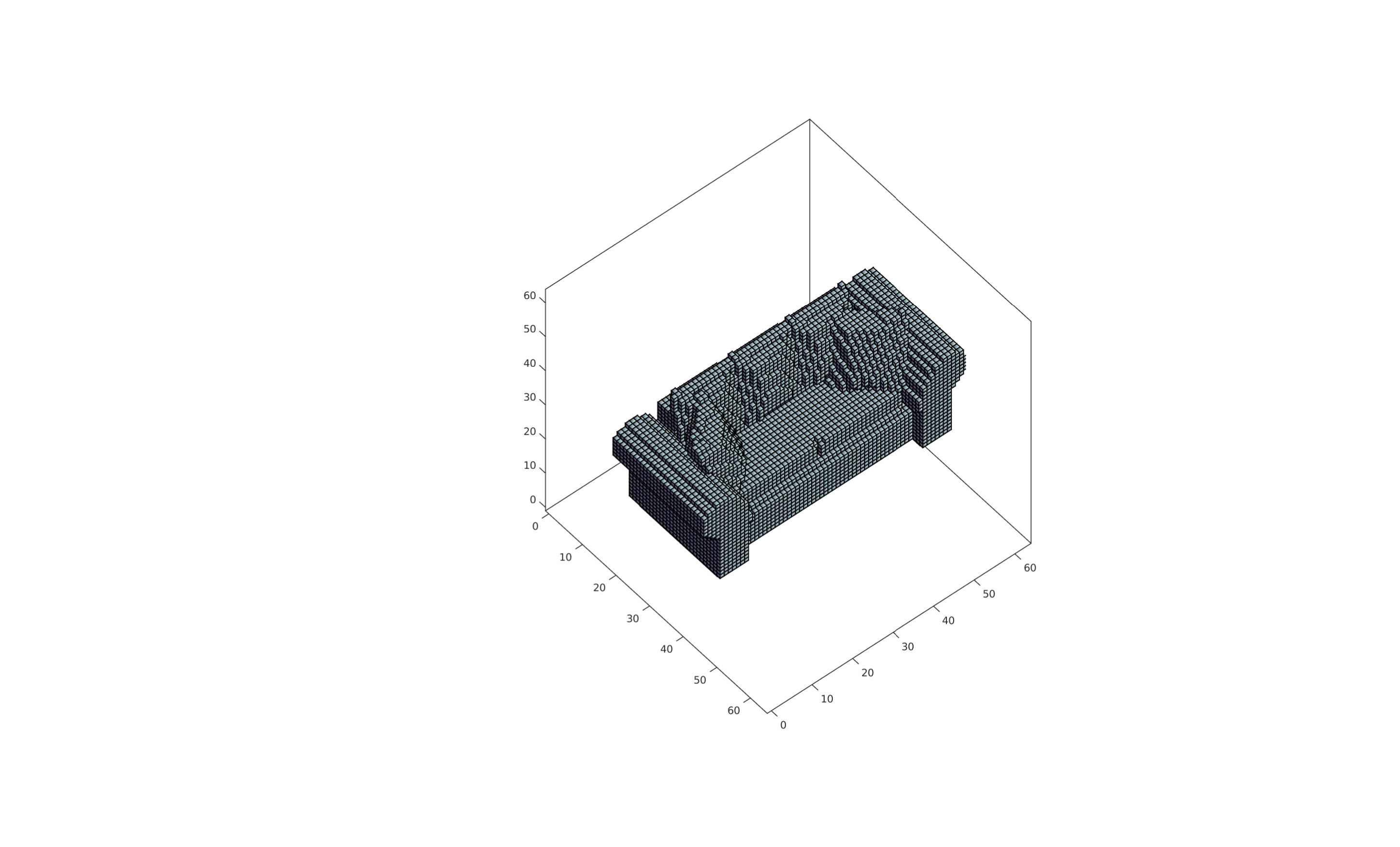} &
		\includegraphics[width=0.4\columnwidth, trim=270 50 350 150,
		clip=true, keepaspectratio]{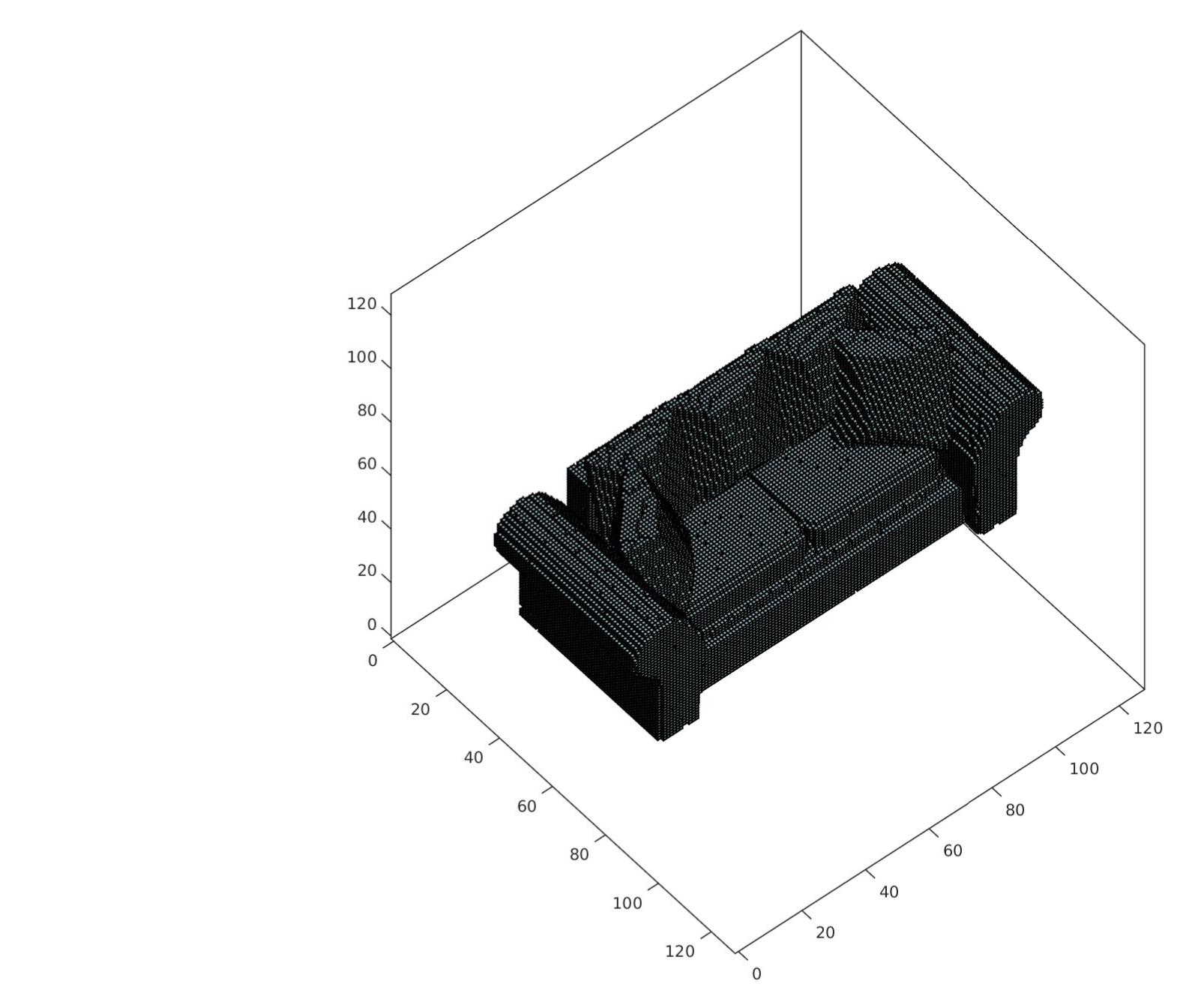} \\
		
		$\;\;\;\;r=64$ &
		$\;\;\;\;r=128$ \\
		
		$\;\;\;\;\overline\rho=5.3\%$ &
		$\;\;\;\;\overline\rho=2.7\%$ \\
	\end{tabular}
	\caption{Sparsity of volumtric 3D data representations. The density
		$\overline\rho$ of occupied voxels is low for 3D models from
		Modelnet40, and decreases with growing voxel resolution $r^3$.}
	\label{fig:modelnet_data}
\end{figure}

However, while naturally suited for image processing, regular grids
are not an ideal representation for sparse data such as line drawings
or irregular 3D point clouds (Fig.~\ref{fig:modelnet_data}). E.g., the
latter are typically acquired with line-of-sight instruments, thus the
large majority of points lies on a small number of $2$D surfaces.
When represented as occupancies in a 3D voxel grid, they therefore
exhibit a high degree of sparsity, as most voxels are empty; while at
the same time 3D data processing with CNNs is challenged by high
memory demands
\citep{brock2017gen,wu20153d,maturana2015voxnet,huang2016point}.

\begin{figure}[b!]
	\centering
	\begin{tabular}{ccc}
		\includegraphics[width=0.14\linewidth]{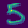} &
		\includegraphics[width=0.14\linewidth]{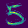} &
		\includegraphics[width=0.14\linewidth]{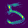} \\
		\includegraphics[width=0.14\linewidth]{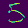} &
		\includegraphics[width=0.14\linewidth]{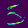} &
		\includegraphics[width=0.14\linewidth]{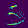}
	\end{tabular}
	\caption{With suitable computational mechanisms, sparsity in the
		input can be preserved throughout the CNN. Shown are the
		activations for one channel of the 1$^\text{st}$, 2$^\text{nd}$
		and 3$^\text{rd}$ convolution layers on MNIST for a dense network
		(top) and for a sparse network with the fraction
		of non-zero elements restricted to at most $\rho_{up} = 15\%$
		(bottom).}
	\label{fig:mnist}
\end{figure}

A conceptually straight-forward counter-measure is to make explicit
the sparsity of the feature maps and store them in a sparse data
format, see Fig.~\ref{fig:mnist}.
Moreover, it can be beneficial to also represent the CNN parameters in
a sparse manner to improve runtime and -- perhaps more important with
modern, deep architectures -- memory footprint; especially if the
sparsity is promoted already during training through appropriate
regularisation.

In a sufficiently sparse setting, a
significant speed-up can be achieved if one performs convolution
\emph{directly}, incrementally updating a layer's output map only
where there are non-zero entries in the input map as well as non-zero
filter weights.
This has recently been confirmed independently by two concurrent works
\citep{Engelcke2017Vote3Deep,JongsooPark2017Faster}.
Computing convolution directly guarantees that only the minimal
necessary amount of operations is carried out. However, performing
selective updates only at indexed locations makes parallelisation
harder.
This may be the reason why, to our knowledge, no practical
implementations with sparse feature maps exist.
In this work we develop a framework to exploit both sparse feature
maps and sparse filter parameters in CNNs. To that end
\begin{enumerate}
	\item we provide sparse versions of the fundamental building blocks of
	CNNs, namely a sparse \emph{Direct Convolution Layer}, as well as
	sparse \emph{ReLU} and \emph{max-pooling} layers;
	\item we extend the back-propagation algorithm to preserve sparsity,
	and make our sparse layers usable with existing optimisation
	routines that are employed in modern deep learning frameworks
	designed for dense data;
	\item we propose to add a density-dependent regulariser that
	encourages sparsity of the feature maps, and a pruning step that
	suppresses small filter weights. This regularisation guarantees
	that the network gets progressively faster at its task, as it
	receives more training.
\end{enumerate}
All these steps have been implemented on GPU (and also on CPU) as
extensions of
\emph{Tensorflow}, for generic $n$-dimensional tensors. The source code
is available at \texttt{https://github.com/TimoHackel/ILA-SCNN}.
In a series of experiments, we show that it outperforms its dense
counterpart in terms of both runtime and memory footprint when
processing sparse data, and makes it possible to process 3D data at
higher voxel resolutions.

The present paper is based on the preliminary version \citep{hac-18}.

\section{Related work}
\label{sec:related_work}

\subsection{Dense CNN for sparse data}
Neural networks offer the possibility to learn a mapping from input
data to a desired output end-to-end, completely avoiding heuristic
feature design and feature selection.
Deep, convolutional neural networks are at present the method of
choice for a wide range of 2D image interpretation tasks, including
object recognition~\citep{he2016deep} and
detection~\citep{ren2015faster}, semantic
segmentation~\citep{chen2018deeplab}, motion~\citep{ilg2017flownet} and
depth estimation~\citep{zhou2017unsupervised}, and many others.
Moreover, deep learning has been adapted to 3D voxel
grids~\citep{prokhorov2010,lai2014unsupervised,maturana2015voxnet,wu20153d},
RGB-D images~\citep{song2015deep} and video~\citep{karpathy2014large}.
Being completely data-driven, these techniques have the ability to
capture appearance as well as geometric object properties. Their
multi-layered, hierarchical architecture is able to capture and encode
a large amount of contextual information.
A serious drawback of the straight-forward generalisation to 3D is
that operating on (dense) voxel grids derived from (originally sparse)
point clouds generates a the huge memory overhead for encoding large
contiguous volumes of empty space%
\footnote{Respectively, occupied space, since the binary occupancy
	label can be flipped. However, in practice situations with only
	little free space are much rarer.} %
Moreover, also computational complexity grows cubically with voxel
grid resolution, but in fact the data is concentrated on few object
surfaces.

\subsection{Data sparsity}
Therefore, more recent 3D-CNNs make use of the \emph{sparsity of
	occupied voxels} prevalent in many practical 3D
datasets. In~\citet{graham2014spatially} a sparse CNN is introduced,
which is however limited to small resolutions (in the paper, up to
$80^3$) because of the rapidly decreasing sparsity due to repeated
convolutions.
Another strategy is to resort to an octree representation, where empty
space (and potentially also large, geometrically simple object parts)
are represented at coarser scales than object
details~\cite{Riegler2017OctNet,tatarchenko2017}. Since the octree
partitioning is a function of the object at hand, an important
question is how to automatically adapt to new, previously unseen
objects at test
time. While~\cite{Riegler2017OctNet,wang2017ocnn}
assume that the octree structure of the output is known at test
time, \cite{tatarchenko2017,riegler2017onfusion,wang2018aocnn} learn
to predict the octree structure together with the labels. This makes
it possible to solve tasks where 3D geometry must be generated,
e.g., fusion of partial 3D reconstructions and shape synthesis by
interpolation.
In~\citet{HaeneHSP17} a coarse-to-fine scheme is used to
hierarchically predict the values of small blocks of voxels in an
octree.
Another strategy is to rely only on a small subset of discriminative
points, while neglecting the large majority of less informative
ones~\citep{Li2016,qi2017,Qi2017b}. The idea is that the network learns
how to select the most informative points and aggregates information
into global descriptors of object shape via fully-connected
layers. This allows for both shape classification and per-point
labeling using only a small subset of points, resulting in significant
speed and memory gains.
Bilateral convolutional layers~\citep{jampani2016learning} map the data
into permutohedral space, thus also exploiting sparsity in the data,
but do not have a mechanism to exploit parameter sparsity.
Recently,~\citet{graham2017submanifold,
	graham2017submanifold_seg} advocate the strategy to perform
convolutions only on non-zero elements in the feature map and find
correspondences with the help of hash tables.
Empirically, it works fairly well to limit activations to non-zero
elements of the input, although in principle it suppresses information
exchange across empty space, thus potentially slowing down and holding
back the learning.

\subsection{Parameter sparsity}
Several works address the situation that the model \emph{parameters}
are sparse.
\citet{denil2013predicting} reduce the network parameters
by exploiting low rank matrix factorisation.
\citet{liu2015sparse} exploit the decomposition of matrices to
perform efficient convolutions with sparse kernel parameters.  Some
authors~\citep{JaderbergVZ14,denton2014exploiting} approximate
convolution filters to achieve a faster runtime, moreover it has
been proposed to reduce the number of parameters by pruning
connections~\citep{han2015learning} or imposing sparsity in an already
trained network~\citep{wen2016learning}.

\subsection{Direct convolutions}
The works
of~\citet{JongsooPark2017Faster,Engelcke2017Vote3Deep,parashar2017scnn}
are the most related ones to our approach, in that they also compute
convolutions in a direct manner to more efficiently convolve sparse
feature maps with sparse filters.
While~\citet{JongsooPark2017Faster} use compressed rows as sparse
format for the filter parameters, neither
\citet{Engelcke2017Vote3Deep} nor \citet{JongsooPark2017Faster} uses a
sparse format for both filter parameters and feature maps.
\citet{parashar2017scnn} implement sparse convolutions
on custom-designed hardware to achieve an energy- and memory-efficient
CNN. Even though all three works follow a similar idea, only
the latter exploits sparsity in both the parameters and the data, with
compressed sparse blocks; but requires dedicated, non-standard
hardware.
Recently, sparse convolution has been extended to
allow for arbitrary kernel shapes by explicitly storing the input
indices for each output, thus including sparse kernel
weights~\citep{choy2019}.
Moreover, in very recent work the use of sparsity has been pushed
further for the particular case of point
clouds:~\citet{boulch2019,thomas2019iccv} evaluate the convolution in
continuous space, but only at the input point locations.

\section{Method}
\label{sec:sparse_layers}

It is a recurring theme in computer science to speed up computations
and reduce memory usage by exploiting sparsity in the processed data.
Here, we propose a number of techniques to do the same for the
specific case of CNN layers, always keeping in mind the specific
architectural requirements and limitations of current GPUs, which are
the prevalent hardware used to run neural networks.
Throughout, sparse tensors are represented and manipulated in a format
similar to \emph{Coordinate List}.%
\footnote{We have also experimented with other sparse formats, like
	compressed sparse blocks; but found none of them to work as well,
	in part due to limitations and idiosyncrasies of current GPU
	hardware.} %
Such a format is available in modern software frameworks for deep
learning (for instance, the "SparseTensor" structure in \emph{Tensorflow}).
Indices of populated voxels in the grid, as well as the corresponding
data values, are stored in separate tensors of equal size.  To
minimise memory overhead, the indices of the form $\{batch, index_x,
index_y, ..., channel\}$ are compressed into unique $1D$ keys and only
expanded when needed.

To achieve coalesced memory access, which permits efficient caching,
the tensors for feature maps are sorted w.r.t.\ batches and within
each batch w.r.t.\ channels. Likewise, filter weights are sorted
w.r.t.\ the output channels and within each channel w.r.t.\ the input
channels.

Compared to dense tensors, the sparse representation inevitably adds
some overhead. In our implementation we use $64$ bit keys, and $32$
bit depth for feature maps. Consequently, storing a dense feature map
(100\% density) requires 3$\times$ more memory in the \emph{Coordinate
	List} format. For densities $<$33\% the sparse representation is
more efficient, and at low densities the savings can be very
significant, e.g., at density 1\% it uses 97\% less memory.
In this context, we note that even less dramatic savings, say $>$50\%,
can have important consequences in practice: a main bottleneck of deep
neural networks, especially for 3D or even 4D data, is the available
on-board memory on the GPU, which limits the batch size during
training and inference. Overly small batches slow down or even harm
the training -- up to the point where training becomes impossible
because there is not enough memory to pass a single example through
the net.

\subsection{Sparse convolution}
Our convolutional layer is designed to work with sparse tensors for
both feature maps and filter weights. Feature maps are updated
incrementally with \emph{atomic operations}, c.f.\ Algorithm
\ref{DSConvWA}.
In that respect it is similar to two concurrent works
\citep{parashar2017scnn,JongsooPark2017Faster}.
Atomic operations are a feature of modern GPUs, designed to be small
enough to be thread-safe, without having to use a locking mechanism.
The incremental update is unfortunately not a perfect match for the
current hardware design, since atomic operations are slightly slower
than non-atomics: existing off-the-shelf GPUs do not offer native
support for atomic floating point operations in shared memory,
although they do for more costly CAS instructions.
Nonetheless, incremental updating is significantly faster, because it
performs only the minimum number of operations necessary to obtain the
convolution result, while avoiding to multiply or add zeros.

\begin{algorithm}[tb]
	\caption{Direct Sparse Convolution with Attention}\label{DSConvWA}
	\label{alg:direct_sparse_conv_forward_pass}
	\begin{algorithmic}[1]
		\State decompress filter and data indices from $1$D to $k$D
		\For{$b \in [0:\,$\emph{batch\_count}$]$}
		\For{$oc \in [0:\,$\emph{out\_channel\_count}$]$}
		\State initialize dense \emph{buffer} with $0$
		\For{$ic \in [0:\,$\emph{in\_channel\_count}$]$)}
		\For{$\{id, val\} \in$ data($b, ic$)}
		\For{$\{fid, fval\} \in \,$\emph{filter($oc,ic$)}}
		\State compute $uid$ with get\_update\_id($id, fid$)
		\State atomically add $val \cdot fval$ to \emph{buffer} at $uid$
		\EndFor
		\EndFor
		\EndFor
		\State get \emph{non-zero} entries from \emph{buffer}
		\State add \emph{bias} to non-zero entries in \emph{buffer}
		\State select $k$ largest responses from \emph{non-zero} entries
		\State compress ids of $k$ largest responses from $k$D to $1$D
		\State write $k$ largest features and ids as sparse output
		\EndFor
		\EndFor
	\end{algorithmic}
\end{algorithm}

The sparse convolution is computed sequentially per output channel and
batch, but in parallel across input channels, features and filter
weights. Its result is stored in a temporary, dense buffer with
batch size and channel depth 1.
This buffer increases quadratically for 2D images, cubically for 3D
volumes, etc.
Still, it is in practice a lot smaller than a dense tensor holding all
channels for the entire batch, such that volumes up to $512^3$ can be
processed on a single commodity graphics card (Nvidia Titan Xp, 12
GB).

\subsection{Preserving sparsity with attention}

\begin{figure}[b]
	\centering
	\subfloat{%
		\includegraphics[height=0.08\textwidth,keepaspectratio]{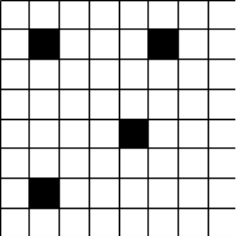}
	}
	\hspace{0.5mm}
	\subfloat{%
		\includegraphics[height=0.08\textwidth,keepaspectratio]{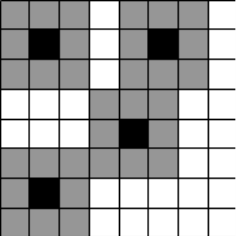}
	}
	\hspace{5mm}
	\subfloat{%
		\includegraphics[height=0.08\textwidth,keepaspectratio]{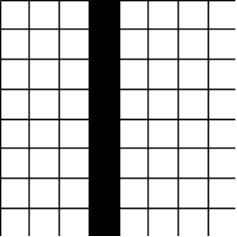}
	}
	\hspace{0.5mm}
	\subfloat{%
		\includegraphics[height=0.08\textwidth,keepaspectratio]{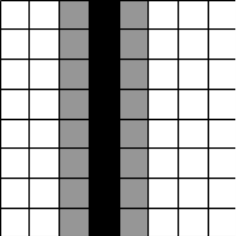}
	}
	\caption{Fill-in (loss of sparsity) due to convolutions
		depends on the data distribution. Uniformly distributed data is
		affected most strongly, e.g., in $3$D every $3 \times 3
		\times 3$ filter will increase the density by a factor of $27$,
		until data is dense.}
	\label{fig:fill-in}
\end{figure}

Convolution with kernels larger than $(1\times 1)$, by construction,
generates fill-in, i.e., it generates an output feature map that is a
lot less sparse than the input. See Fig.~\ref{fig:fill-in}.
In the presence of well-spaced sparse structures, like for example the
surfaces in a 3D scan, this ``smearing out'' of the sparse inputs into
empty space only has a small influence on the output of the network (see the 
experiments). But it considerably increases memory consumption
and runtime, especially when occurring repeatedly over multiple
layers.
Intuitively, a stack of convolutional layers ``diffuses'' information
over its cumulative receptive field, but with small convolution
kernels even a deep stack cannot bridge large empty regions and capture
context across them.

Our goal here is to prevent fill-in and maintain a desired level of
sparsity, by dropping small feature responses. To that end, we run a
$k$-selection filter \citep{alabi2012kselection} on each output channel
and keep only the $k$ strongest (non-zero) responses.
The parameter $k$ controls the sparsity of the convolutional layers.
Note, for a given network architecture an upper bound on the number of
non-zero entries directly translates to a bound on the memory
footprint, and for a given hardware configuration also a bound on the
runtime.
The $k$-selection can be seen as an approximation of the exact
convolution. Small responses are suppressed, but with an adaptive
threshold that suppresses only as many values as necessary to uphold
the desired degree of sparsity.
The selection can be interpreted as a form of \emph{attention}, in the
sense that its aim is to manage a limited pool of computational
resources and assign them as needed for the solution of the task.

We have implemented two versions of this simple attention mechanism:
\begin{itemize}
	\item
	The first one acts on the raw convolution result, so it prefers
	large positive responses, making it similar to a rectified linear
	unit;
	\item The second one picks the $k$ responses with the largest
	absolute values, giving preference to responses with large
	magnitude.
\end{itemize}
Empirically the first version works better, see
Section~\ref{sec:evaluation}.

The overall time complexity to convolve data of dimension $k$,
resolution $s_d$ and density $\rho_d$, with filters of size $s_f$ and
density $\rho_f$, is
\begin{equation}
O\big( (\rho_d \cdot \rho_f \cdot s_f^k \cdot c_{in} + log(s_d^k) ) \cdot
s_d^k \cdot c_{out} \cdot b\big)\;,
\end{equation}
with $b$ the number of batches and $c_{in},c_{out}$, the number of
input and output channels, respectively.

\begin{algorithm}[tb]
	\caption{Backpropagation for convolutional layer}\label{euclid}
	\label{alg:direct_sparse_backprop}
	\begin{algorithmic}[1]
		\State initialize \emph{bp\_data} with shape(\emph{input\_values}) and $0$
		\State initialize \emph{bp\_filter} with shape(\emph{filter\_weights}) and 
		$0$
		\State decompress filter and data indices from $1$D to $k$D
		\For{$b \in [0:\,$\emph{batch\_count}$]$}
		\For{$oc \in [0:\,$\emph{out\_channel\_count}$]$}
		\State initialize dense \emph{buffer} with gradients($b, oc$)
		\For{$ic \in [0:\,$\emph{in\_channel\_count}$]$)}
		\For{$\{id, val\} \in$ data($b, ic$)}
		\For{$\{fid, fval\} \in \,$\emph{filter($oc,ic$)}}
		\State compute $uid$ with get\_update\_id($id, fid$)
		\State get gradient $g$ from \emph{buffer} at $uid$
		\State atomically add $g \cdot fval$ to \emph{bp\_data} at $id$
		\State atomically add $g \cdot val$ to \emph{bp\_filter} at $fid$
		\EndFor
		\EndFor
		\EndFor
		\EndFor
		\EndFor
	\end{algorithmic}
\end{algorithm}

\subsection{Pooling layer}
Our sparse pooling layer has three straight-forward stages.
First, assign features to an output (hyper-)voxel, by dividing
the data channels of their index by strides.
Second, sort the data w.r.t.\ voxels, so that responses
within the same voxel are clustered together.
Third, apply the pooling operator separately to each
cluster. The time complexity for this is
\begin{equation}
O\big(\rho_d \cdot s_d^k \cdot log(\rho_d \cdot s_d^k)\cdot c_{in} \cdot b\big)\;.
\end{equation}

\subsection{Direct sparse backpropagation}
Our target for back-propagation is again to avoid carrying out
operations that are not needed due to sparsity.
Clearly, we must propagate error gradients only to those features
which have produced evidence, in the form of non-zero responses during
the forward pass.

Yet, the problem of rapid fill-in, already discussed for the forward
pass, equally affects the backward pass: back-propagation through a
convolution layer is itself a convolution that spreads out non-zero
values over a neighbourhood, thus increasing memory demands and
runtime.
Contrary to the forward pass, it is not advisable to bound the fill-in
with $k$-selection, as this could seriously slow down or even impair
the training: during back-propagation, large error gradients can flow to
zero activations and vanish, whereas smaller gradients flowing towards
non-zero activations might be missed due to the selection.
Hence, we propose to use a stricter back-propagation, which only
propagates errors $L$ to non-zero features $x$ and model parameters
$w$:
\begin{equation}
\begin{split}
&\frac{\partial L}{\partial x_i} =
\left\{
\begin{array}{ll}
0 & \text{for } x_i = 0\\
\frac{\partial L}{\partial y}
\frac{\partial y}{\partial x_i},
& \text{else}
\end{array}
\right.
\\
&\frac{\partial L}{\partial w_i} =
\left\{
\begin{array}{ll}
0 & \text{for } w_i = 0\\
\frac{\partial L}{\partial y}
\frac{\partial y}{\partial w_i},
& \text{else}
\end{array}\right.\;.
\end{split}
\end{equation}
Here, weights are considered equal to zero only if they have been
explicitly removed by pruning, so as to avoid suppressing the gradients
of weights that pass through $w_i\!=\!0$ while changing sign.
Note the structural similarity of our approximated back-propagation to
back-propagation through any layer with $ReLU$ activation:
Conventional back-propagation sets values to zero as a function of the
layer output $y_i$, whereas our scheme sets them to zero as a function
of the input $x_i$.

Neglecting zero-elements slightly reduces the efficiency per learning
iteration, since not all error gradients are propagated
anymore. However, this is offset by several of advantages:
\begin{enumerate}
	\item The tensors used for back-propagation have fixed size and
	shape. Therefore, one can work with fixed and known array
	dimensions, and use optimisation frameworks that have been
	designed for dense data;
	\item By considering only gradients on non-zero elements of the
	forward pass, back-propagation can be implemented in a clean and
	transparent manner. E.g.\, for convolutional layers one obtains
	Algorithm~\ref{alg:direct_sparse_backprop}, which is very similar
	to Algorithm~\ref{alg:direct_sparse_conv_forward_pass};
	\item Once a filter weight has been set to zero, it will remain
	zero. Below, we will describe how this property can be used to
	guarantee that the network gets progressively faster at its task
	as the learning proceeds and it sees more training data.
\end{enumerate}

\subsection{Adaptive density regularisation}
The $k$-selection filter adds to the computational cost of the overall
method, but there is also a computationally more efficient way to
reduce the number of non-zero feature responses. Although that trick,
described in the following, cannot guarantee a bound on the sparsity,
it helps to stay below the bound, such that the more costly
$k$-selection often need not be invoked.

The $ReLU$ non-linearity used in most modern CNNs, by definition,
truncates negative activations to zero while leaving positive ones
unchanged. This means that we can encourage sparsity by lowering the
values (not magnitudes) of filter weights and biases.
In doing so, more weights will drop below $0$ and will be extinguished
by the subsequent $ReLU$.

Going back to the idea of maintaining an \emph{optimal}, rather than
\emph{maximal}, level of sparsity, one can use the same idea to reduce
sparsity when there are fewer non-zero values than would fit into the
memory and computation budget.
When too many activations drop below $0$, one simply drives the filter
parameters up, so that fewer responses are suppressed by the $ReLU$.
To achieve the desired effect, we simply add a bias $b$ to the
$L_2$-regulariser on the weights, so that the regulariser becomes
$\sum(w+b)^2$.  The scalar $b$ is positive when the density $\rho$ is
too large, and negative when it is overly small:
\begin{equation}
b =
\begin{cases}
o + b_1 \cdot (\rho - \rho_{up}) & \text{if } \rho > \rho_{up}\\
&\text{(exceeds available resources)}\\
-b_2 \cdot (\rho_{up} - \rho) & \text{if } \rho \le \rho_{up}\\
& \text{(not using all resources)}
\end{cases}
\end{equation}
with $\rho_{up}$ the upper bound implied by the $k$-selection filter,
and $o$, $b_1$, $b_2 \ge 0$ control parameters.  The bias is
asymmetric and adds an extra penalty $o$ for exceeding the available
resources, since that would trigger the $k$-selection filter and cause
additional computational load.

\subsection{Parameter pruning}
As explained above, our
training algorithm has the following useful properties: \emph{(i)} The
regulariser encourages small model parameters.  \emph{(ii)} The sparse
back-propagation ensures that, once set to zero, model parameters do
not reappear in later training steps.
Together, these two suggest an easily controllable way to
progressively favour sparsity during training, by one-warning-shot
pruning of weights:
At the end of every training epoch we screen the network for weights
$w$ that are very small, $ |w_i| < \epsilon$.
If the magnitude of a weight $w_i$ stays low for two consecutive
epochs (meaning that it was already close to zero before, and still is
after one epoch of training) we conclude that it has little influence
on the network output and remove it.
We note that a small weight should not be pruned when first detected,
without warning shot: it could have a large gradient and just
happen to be at its zero-crossing from a large positive to a large
negative value (or vice versa) at the end of the epoch.
On the contrary, it is less likely to observe a
weight exactly at its zero-crossing twice in a row.

Since such a weight, once set to zero, will remain zero due to the
proposed sparse back-propagation, every pruning can only reduce the
number of non-zero weights.
It is therefore guaranteed that the network become sparser, and
therefore also faster at the task it is learning, as it sees more
training data; until a configuration has been reached that is
``optimal'' in the sense that all weights have significant magnitude
and no further pruning is possible.
We note, it is well documented that biological systems get faster at
cognitive tasks with longer
training~\citep{robertson2007serial,nissen1987attentional}; but we do
not claim that the underlying mechanisms are similar.

\section{Evaluation}
\label{sec:evaluation}

In this section we evaluate the impact of density upper bounds and
regularisation on runtime and classification accuracy.  The sparse
network structures were implemented into the Tensorflow framework and
programmed in C++/CUDA with a python interface. Our experiments were
run on PCs with Intel Core i7 7700K processors, 64GB RAM and Titan Xp
GPUs.
Detailed specifications about the different CNN variants used in the
experiments (in both sparse and dense versions) are given in the
appendix.

To start with, we use a synthetic dataset of sparse random tensors to
evaluate the memory footprint and runtime of our convolutional layer,
and to compare it against the dense layers of Tensorflow version 1.4
(compiled with Cuda 9.0 and CuDNN 6.0).
We conduct different experiments to evaluate the effects of the sparse
network variant on classification accuracy: First, the impact of upper
bounds on classification is evaluated by performing a grid search on
the upper bound $\rho_{up}$ in the convolutional layers.  For this
experiment, a thresholded version of the MNIST data set
\citep{lecun1998gradient} is used, as it is small enough to perform
grid search in a reasonable amount of time and can be interpreted as
sparse data (1D lines in 2D images).  Second, the effects of pruning
on runtime and classification accuracy are shown using the Modelnet
data set \citep{wu20153d}, by varying the strength $\lambda$ of the
regularisation.  Modelnet40 provides 3D CAD models of $40$ different
classes.  Furthermore, the classification results of different
baseline methods are compared on this data set, as well as the more
recent ScanNet~\citep{dai2017scannet}. We train with
$adagrad$~\citep{duchi2011adaptive} with learning rate
$0.001$. Convergence (change of loss $<10^{-5}$) takes approximately
100 epochs.

\begin{figure*}[t]
	\centering
	\begin{tabular}{ccc}
		\includegraphics[height=0.25\textwidth,keepaspectratio]{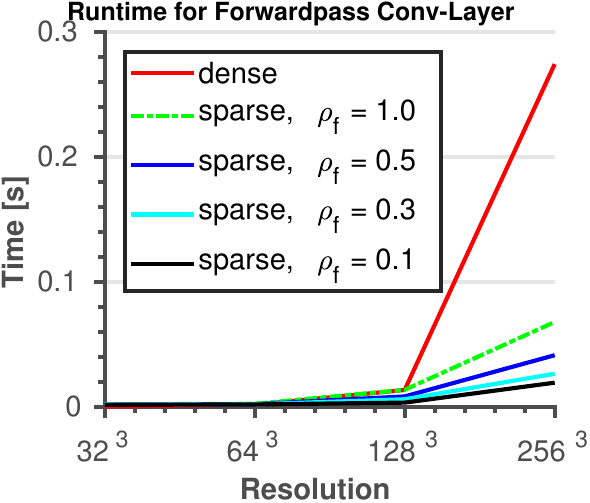}
		& $\quad$ &
		\includegraphics[height=0.25\textwidth,keepaspectratio]{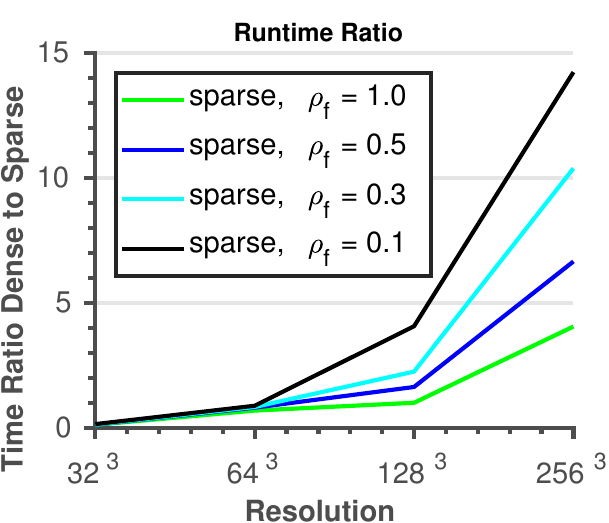}
	\end{tabular}
	\caption{Runtime [s] of a dense convolution layer in Tensorflow
		and of our sparse convolution layer, for random tensors of
		different resolutions $r^3$. At high resolutions the sparse
		version is much more efficient.}
	\label{fig:runtime_conv_layer}
\end{figure*}

\begin{table*}
	\centering
	\begin{tabular}{|c|c|c|c|c|c|}
		\hline
		Resolution & $32^3$ & $64^3$ & $128^3$ & $256^3$ & $512^3$\\
		\hline
		Dense [GB]  & $0.04$ & $0.27$ & $2.15$ & $17.18$ & $137.28$\\
		Sparse 32bit [GB] & $2 \cdot 10^{-3}$ & $8 \cdot 10^{-3}$ & $0.03$ & --- & 
		---\\
		Sparse 64bit [GB] & $3 \cdot 10^{-3}$ & $0.013$ & $0.05$ & $0.2$  & 
		$0.8$\\
		Sparse Temp [GB] & $3 \cdot 10^{-4}$ & $0.002$ & $0.016$ & $0.13$ & 
		$1.07$\\
		\hline
	\end{tabular} 
	\caption{Memory consumption of a dense conv layer in Tensorflow
		and of our sparse conv layer, for different resolutions $r^3$,
		with $\rho_{up}=1/r$, minibatch size $32$ and output depth
		$8$. At high resolutions the sparse version is much more
		efficient.}
	\label{tab:memory_footprint}
\end{table*}

\begin{table*}
	\centering
	\begin{tabular}{|c|c|c|c|c|c|}
		\hline
		Resolution & $16^3$ & $32^3$ & $64^3$ & $128^3$ & $256^3$\\
		\hline
		Octnet &  $16\text{s}$ &  $1\text{m}6\text{s}$ & 
		$5\text{m}30\text{s}$ & 
		$34\text{m}43\text{s}$ & --- \\
		& $3.83$ MB & $20.49$ MB & $114.23$ MB & $549.28$ MB & --- 
		\\
		\hline
		Sparse & $2\text{m}10\text{s}$ & $5\text{m}28\text{s}$ & 
		$11\text{m}36\text{s}$ 
		& $6\text{m}20\text{s}$  & $21\text{m}11\text{s}$ \\
		& $11.25$ MB & 45 MB & 180 MB & 720 MB 
		& 2880 MB \\
		\hline
	\end{tabular} 
	\caption{Training time and memory footprint
		of one epoch on Modelnet40 at different resolutions $r^3$,
		for OctNet and for the corresponding version of our sparse
		network.  In all cases we set $\rho_{up}=1/r$ and use
		minibatch size 32 for single epoch. At $256^3$ Octnet runs out of 
		memory.}
	\label{tab:octnet_runtime}
\end{table*}


\subsection{Runtime and Memory Footprint}
\begin{figure}[t]
	\centering
	\includegraphics[height=0.25\textwidth,width=0.3\textwidth]{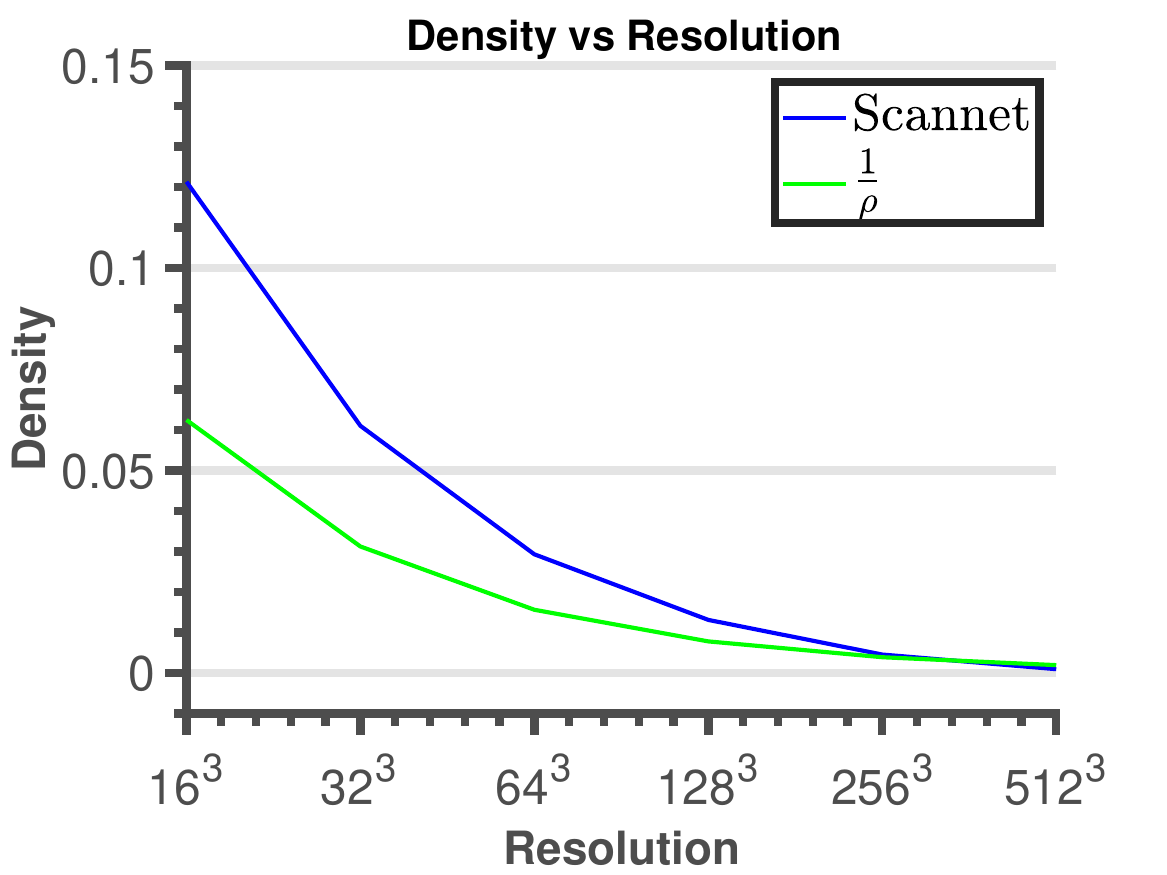}
	\caption{Voxel grid density
		vs.\ resolution. At very low resolutions the density is up
		to $\frac{2}{r}$, but with increasing resolution it
		quickly decays towards $\frac{1}{r}$, and in fact even
		falls slightly below that idealised value at resolution $512^3$.}
	\label{fig:density}
\end{figure}

For the evaluation of runtime, convolutions are performed on a sparse
voxel grid filled with random numbers.  The resolution of the voxel
grid $r^3$ is varied between $r=16$ and $r=256$.  To achieve the
expected data density of a $2$D surface in a $3$D voxel grid, the data
density $\rho$ as well as the upper bound on the per-channel density
$\rho_{up}$ are set to $\rho = \rho_{up} = \frac{1}{r}$.
Empirically, the density of real datasets like
Scannet is indeed $\approx\frac{1}{r}$ at high resolutions that call
for sparse methods. At very low resolution it is a bit higher
($\frac{1.5}{r}$ to $\frac{2}{r}$) due to the presence of multiple
surfaces, see Figure~\ref{fig:density}.

To run dense convolution at resolution $r=256$, the mini-batch size
and channel depth had to be set to $1$ (Protobuf limits each single
tensor to $2$ GB), while the number of output channels was set to
$8$.
The density $\rho_f$ of the filter weights is varied between $0.1$ and
$1$.
As baseline we use the convolutional layer of standard
Tensorflow~\citep{abadi2016tensorflow}, which performs convolutions via
the fast Fourier transform and batched general matrix-matrix
multiplication from cuBlas, as front end to
cuDNN~\citep{chetlur2014cudnn}.
We note that processing only a single input channel does not play to
the strength of our sparse network.
Moreover, Tensorflow is able to use the full capability of the GPU,
while our implementation is limited to operating in global memory, due
to the weak support for atomic floating point operations in shared
memory.  The effect of this limitation is particularly pronounced at
small resolution and high density, whereas for high resolutions and
low densities its influence fades. In particular, at $r = 256$ and
$\rho_f = 0.1$, we are $14\times$ faster with strong density
regularisation, so that the $k$-selection step is bypassed; and still
$7\times$ faster including $k$-selection filtering. See
Figure~\ref{fig:runtime_conv_layer}.

Table~\ref{tab:memory_footprint} shows the memory requirements for
dense and sparse convolution layers at various resolutions~$r$.
Dense convolutions require only a single output tensor. The sparse
implementation uses tensors for indices and data as well as a
temporary buffer, which can be reused in all layers. For the
experiment the data type is $32$bit floating point, for the indices we
consider both $32$bit and $64$bit.%
\footnote{$32$bit indices can only be used for
	resolutions $r\leq 128^3$ due to buffer overflows.} %
As expected, our sparse representation needs less memory at the
sparsity levels of realistic 3D point cloud data. In particular, our
sparse version makes it possible to work with large resolutions up to
$r=512^3$, which is impossible with the dense version on existing
hardware.
Moreover, we also compare both computation time and
memory footprint to Octnet on real Modelnet40 data. Explicit
sparsity is approximately on the same level of memory-efficiency, but 
slower at low
resolutions (that can in many cases even be handled with dense
voxels). At high resolution it clearly outperforms Octnet. In
particular, it still works at resolutions up to $256^3$, where
Octnet runs out of memory (presumably due to higher peak consumption
while building the octree.

\subsection{Contribution of small feature responses}
In the context of sparsity the question arises, whether zero-valued
features contribute valuable information.  Two recent works tried to
answer this question.  On the one hand, \citet{graham2017submanifold}
found that they reach the same accuracies as dense networks for their
application, while completely neglecting zero-valued features.  On the
other hand, \citet{uhrig2017sparsity} concluded that for certain tasks
zero-valued features may be beneficial.
In our framework it is possible to assess the importance of small
feature responses (not limited to exact zeroes) by training neural
networks with varying upper bounds.
For this experiment, CNNs are trained on MNIST without regularisation,
using the $adagrad$ optimiser and a learning rate of $0.01$.
The pixels in MNIST were set to zero when their value $v \in [0,255]$
was below a threshold of $v < 50$, to obtain a sparse dataset with
average input density $\rho_{in} = 0.23$, while the
upper bound $\rho_{up}$ ranges from $\rho_{up}=0.035$ to
$\rho_{up}=0.075$. Note that even though letters can be interpreted as
1D lines in 2D images, the MNIST data has a low resolution of only $28
\times 28$ pixels.  Hence, the data is still not very sparse.
Results after 10 and after 60 training epochs are
given in Table~\ref{tab:mnist_rho}.
Low upper bounds guarantee a small memory footprint, and also yield
slightly faster runtime per epoch. The price to pay is slower
convergence, because some gradients are lost during back-propagation;
and a slight performance penalty for very strict bounds ($<1\%$ for
the strictest setting $\rho_{up}=0.035$).

\begin{table*}
	\centering
	\begin{tabular}{|c|c|c|c|}
		\hline
		$\rho_{21}$ & $0.035$ & $0.05$ & $0.075$\\
		\hline
		Sparse, 10 epochs & $0.954$ & $0.971$ & $0.976$
		\\
		Sparse, 60 epochs & $0.985$ & $0.990$ & $0.992$
		\\
		\hline
	\end{tabular} 
	\caption{Influence of small
		responses. Performance on MNIST with different upper
		bounds on the density of activation maps.}
	\label{tab:mnist_rho}
\end{table*}

\begin{figure*}[t!]
	\centering
	\subfloat{%
		\includegraphics[height=0.25\textwidth,width=0.3\textwidth]{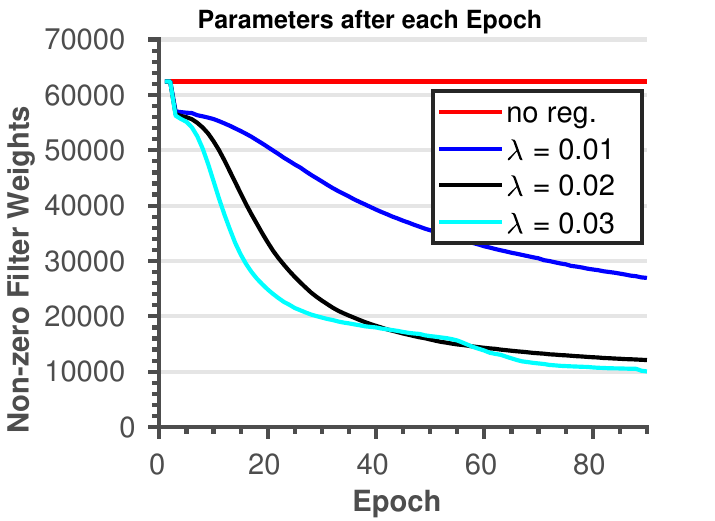}
	}
	\subfloat{%
		\includegraphics[height=0.25\textwidth,width=0.3\textwidth]{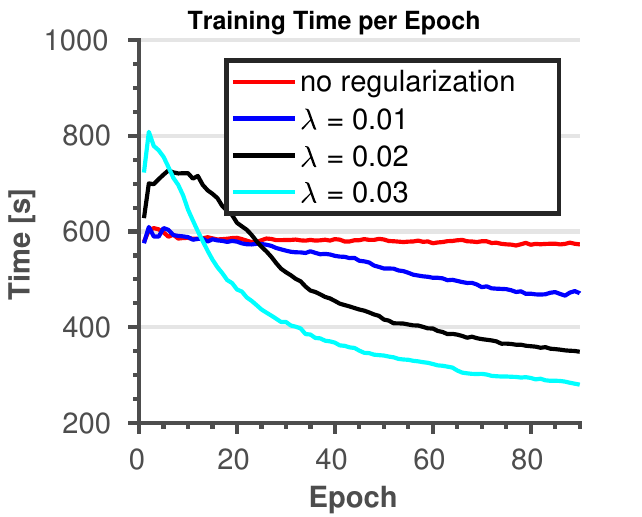}
	}
	\subfloat{%
		\includegraphics[height=0.25\textwidth,width=0.3\textwidth]{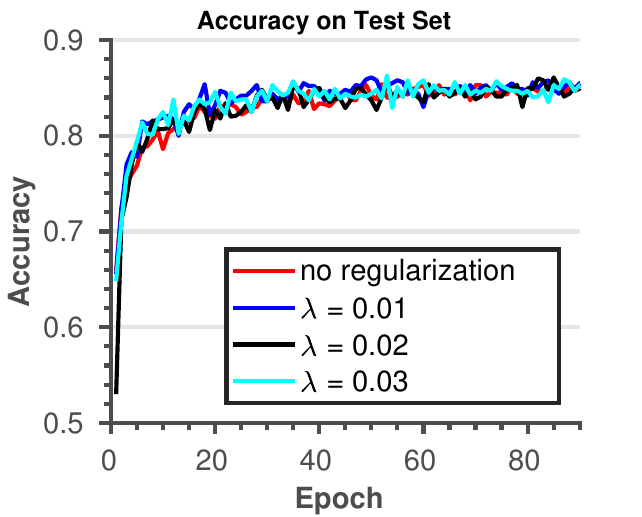}
	}
	\caption{Influence of adaptive density regularisation and pruning
		on \emph{(left)} the number of non-zero filter weights,
		\emph{(middle)} the runtime per training epoch, and
		\emph{(right)} the accuracy on the Modelnet40 test set. Strong
		regularisation and pruning save a lot of memory and time without
		noticeable impact on accuracy.}
	\label{fig:pruning}
\end{figure*}

\begin{figure*}[t!]
	\centering
	\subfloat{%
		\includegraphics[height=0.22\textwidth,keepaspectratio]{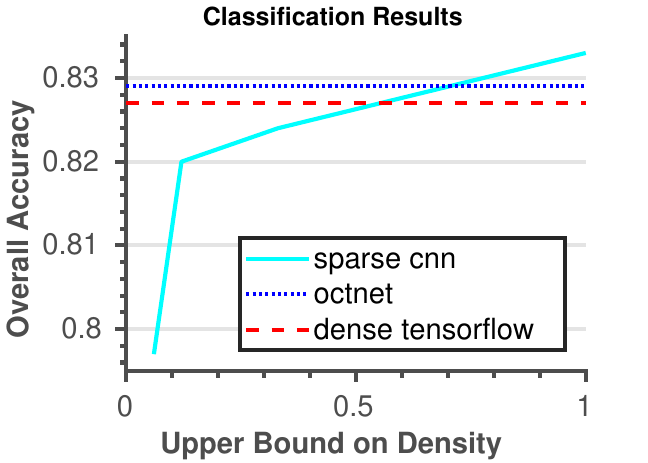}
	}
	\hspace{1mm}
	\subfloat{%
		\includegraphics[height=0.22\textwidth,keepaspectratio]{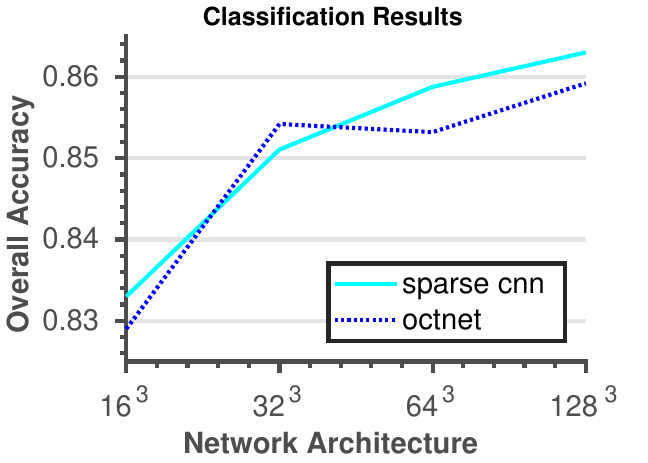}
	}
	\hspace{1mm}
	\subfloat{%
		\includegraphics[height=0.22\textwidth,keepaspectratio]{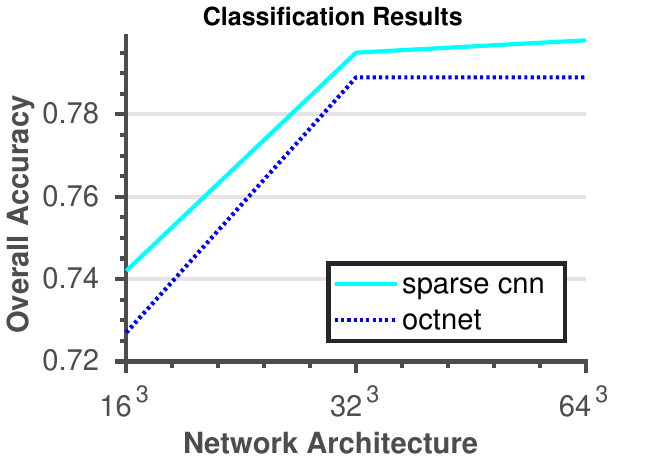}
	}
	\caption{Performance of sparse net, compared to the equivalent
		dense net and Octnet on Modelnet40 \emph{(left, middle)} and
		ScanNet \emph{(right)}. \emph{(left)} accuracy for different
		upper density bounds; \emph{(middle, right)} accuracy vs.\ input
		resolution.}
	\label{fig:modelnet40}
\end{figure*}

\subsection{Regularisation and pruning}
With our sparsity-inducing pruning and regularisation, we expect
faster runtime.  In order to verify this behaviour, neural networks
are trained on Modelnet40 with varying regularisation scales $\lambda
\in \{0, 0.1, 0.2, 0.3\}$.  The bias for density-based regularisation
is computed with $b_1 = b_2 = o = 0.1$.  Stronger regularisation
decimates the number of (non-zero) filter weights faster, as shown in
Figure \ref{fig:pruning}.  It can also be seen that the number of
parameters converges when only important weights are left. The drop in
non-zero weights also reduces runtime.
After $90$ epochs, a network regularised with $\lambda =
0.3$ is $51\%$ faster than one trained without regularisation and
pruning, even though only the first nine out of twelve convolution
layers are set to be sparse.  Strong regularisation
initially causes an increase in runtime, by
driving up the number of non-zero weights to use the available
resources via the bias term $b_2$.
The classification accuracy for all tested regularisation scales
quickly converges to practically identical values, as shown in Figure
\ref{fig:pruning}. We point out that pruning finds the most
suitable sparsity pattern \emph{for a given training set}. When
using a pruned model for transfer learning, it may be safer to
re-initialize the removed filter weights of the sparse representation
with zeros before fine-tuning.

For completeness, we also empirically compare the two
described variants of the $k$-selection filter, where either the $k$
largest signed weights or the $k$ weights with highest magnitude are
preserved. See Figure~\ref{fig:pruning}. The test, run on
Modelnet40, shows that sorting signed weights is preferable, which
is not surprising, given the proven performance of the $ReLU$
activation.

\begin{figure}[t]
	\centering
	\includegraphics[height=0.25\textwidth,width=0.3\textwidth]{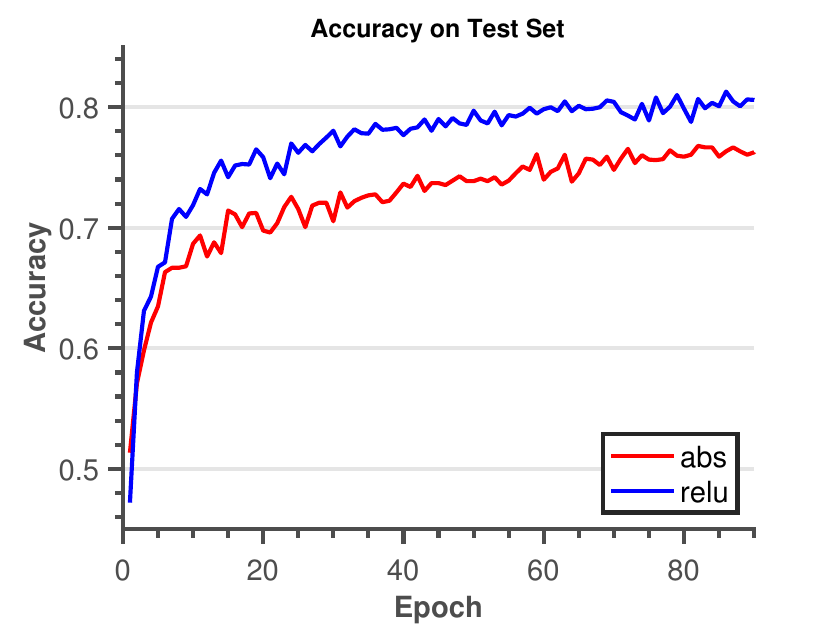}
	\caption{Pruning by signed ($RELU$) or
		unsigned ($ABS$) magnitude. As expected, keeping the
		largest \emph{positive} activations, as in the $ReLU$
		activation, works better. The experiment was run on
		Modelnet40 with resolution $16^3$.}
	\label{fig:k_filt}
\end{figure}

\subsection{Classification performance on Modelnet40}

We compare our upper-bounded neural net and modified back-propagation
against a conventional dense net. To that end we run
the Octnet3 variant of
Octnet~\citep{Riegler2017OctNet}, a dense network without octree
structure, and a sparse version of the same network on Modelnet40, see
Figure \ref{fig:modelnet40}.

First, the input resolution is set to $r = 16^3$, while the upper bound
on the density is varied between $\rho_{in} \in \{0.06, 0.12, 0.33,
1.0\}$.
Both, the conventional dense network and Octnet converge to a
similar overall accuracy of $\approx 0.83$.
For a trivial upper bound $\rho_{in} = 1.0$ the overall accuracy of
our sparse network is also practically the same. Very low upper bounds
up to $\rho_{in} = 0.12$ yield slightly worse results on the $16^3$
inputs, for the lowest bound $\rho_{in} = 0.06$ the drop in
performance reaches $\approx 3$ percent points.
Second, the resolution of the input is gradually increased:
$r~\in~\{16^3,~32^3,~64^3,~128^3\}$.  Both the sparse network and
Octnet yield similar results, for all resolutions.  Octnet performs
slightly better on $r~=~32^3$, while our bounded, sparse network has a
small advantage at all other resolutions.
The two experiments suggest that reasonable upper bounds and
our sparse backpropagation do not reduce significantly
classification accuracy.
We have confirmed this further by running experiments
at different resolutions, see Figure \ref{fig:modelnet40}. For these
experiments we always use our default density bounds as given in the
appendix. The performance of our network closely follows that of
Octnet across a range of voxel sizes.

\begin{figure*}[t]
	\centering
	\begin{tabular}{cccccc}
		\includegraphics[width=0.3\columnwidth]{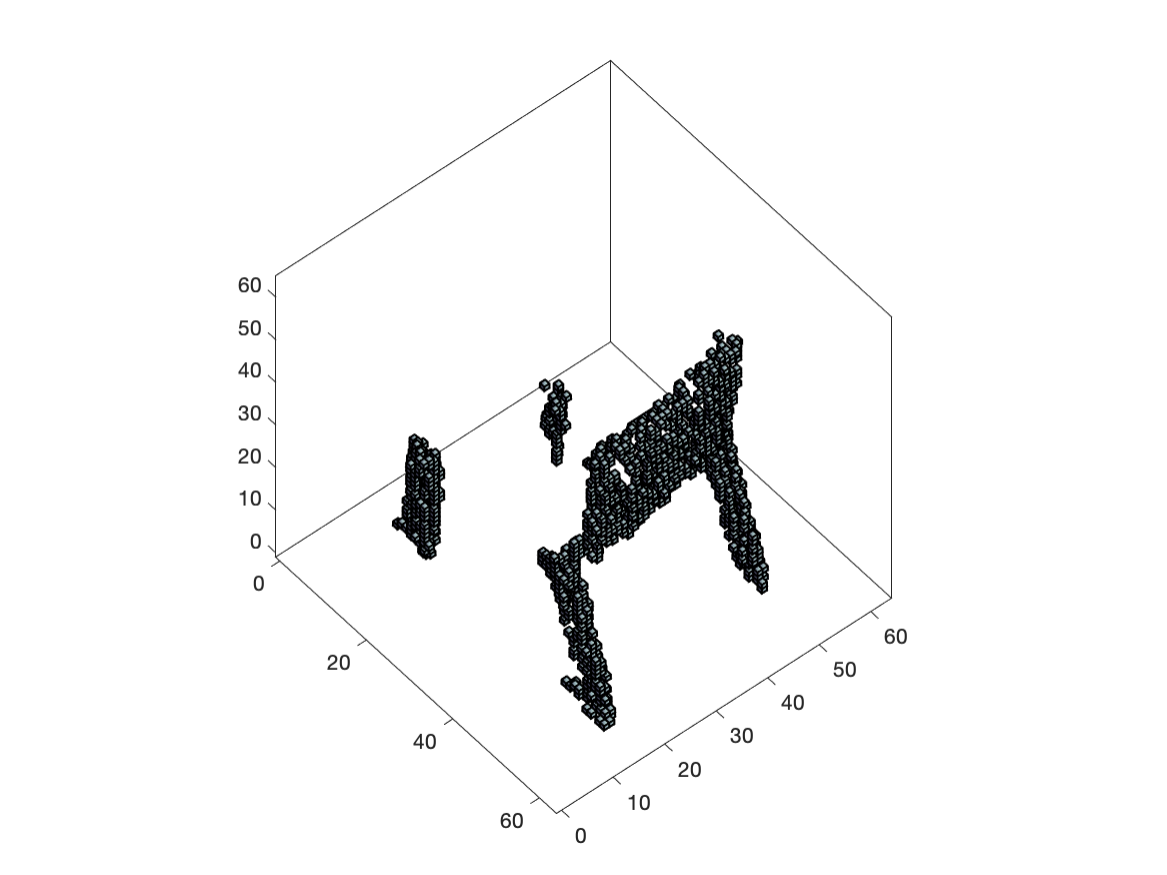} &
		\includegraphics[width=0.3\columnwidth]{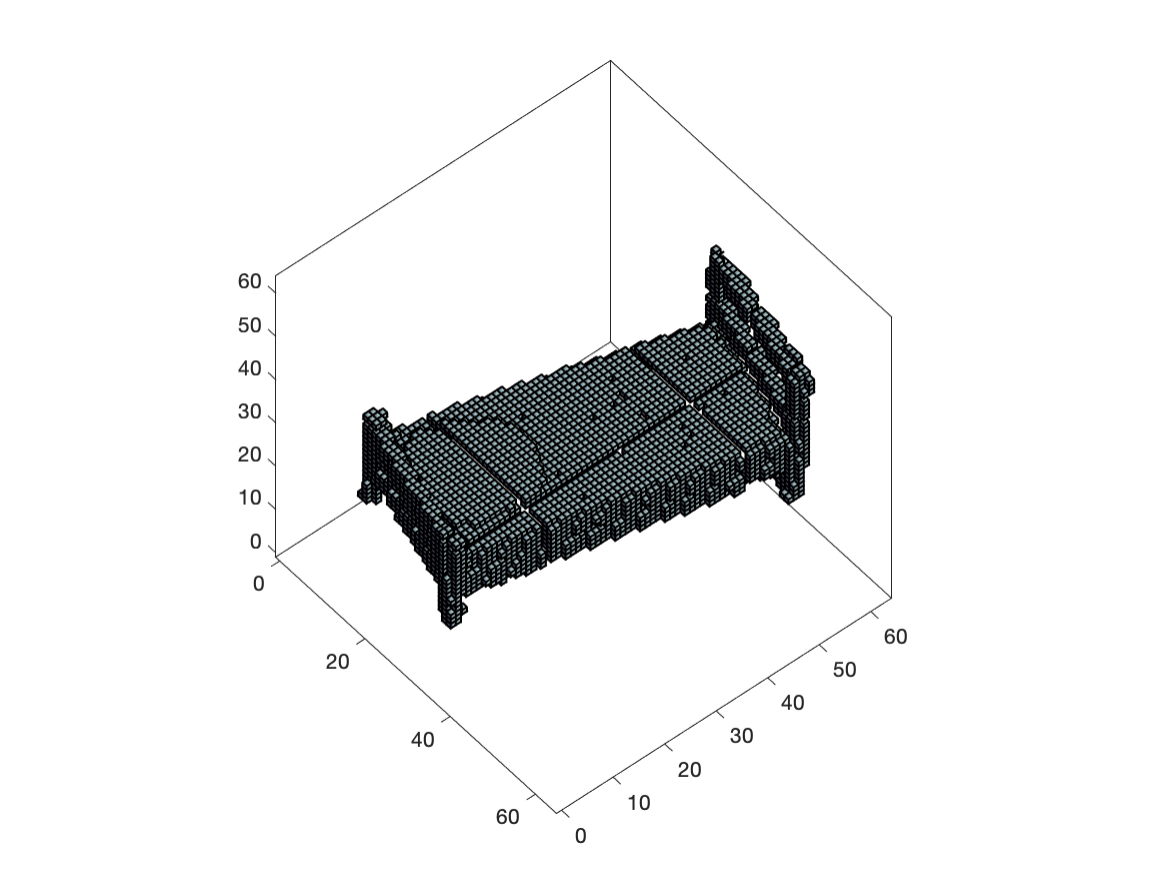}&
		\includegraphics[width=0.3\columnwidth]{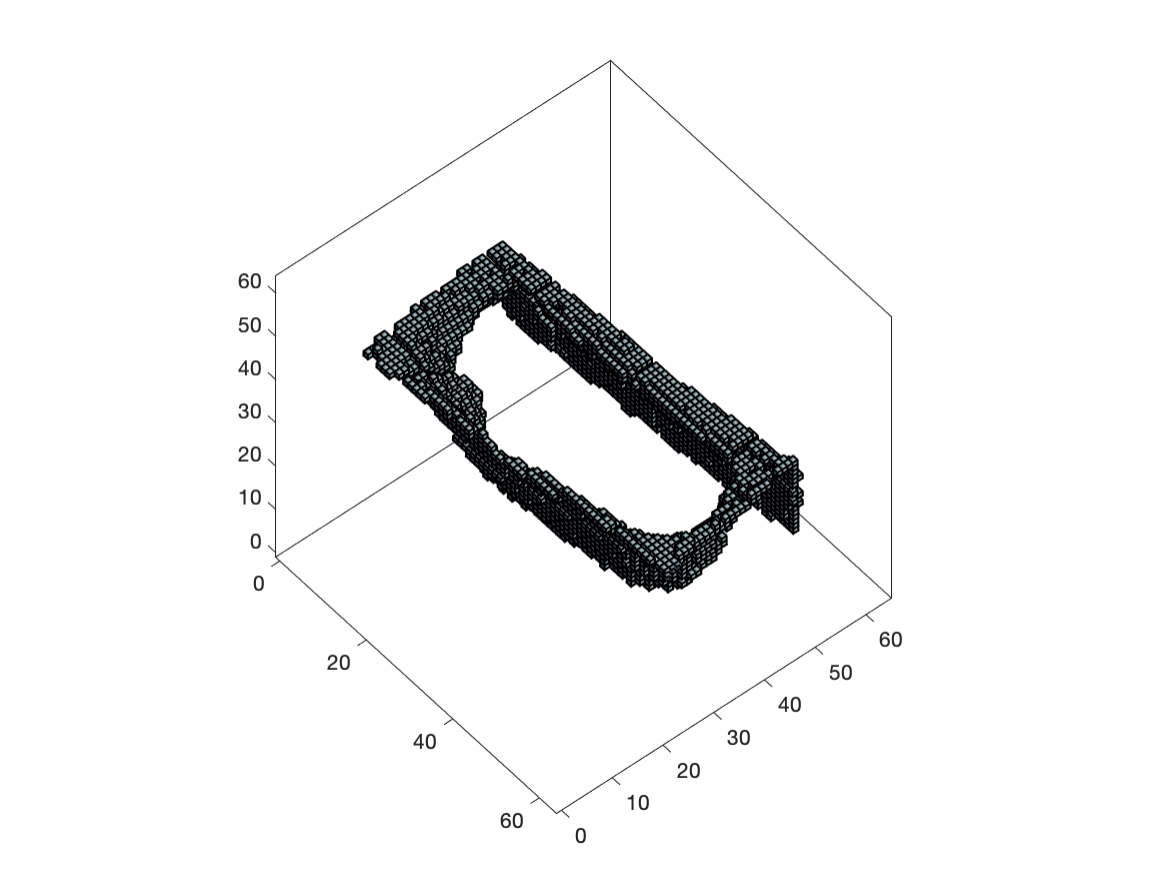}&
		\includegraphics[width=0.3\columnwidth]{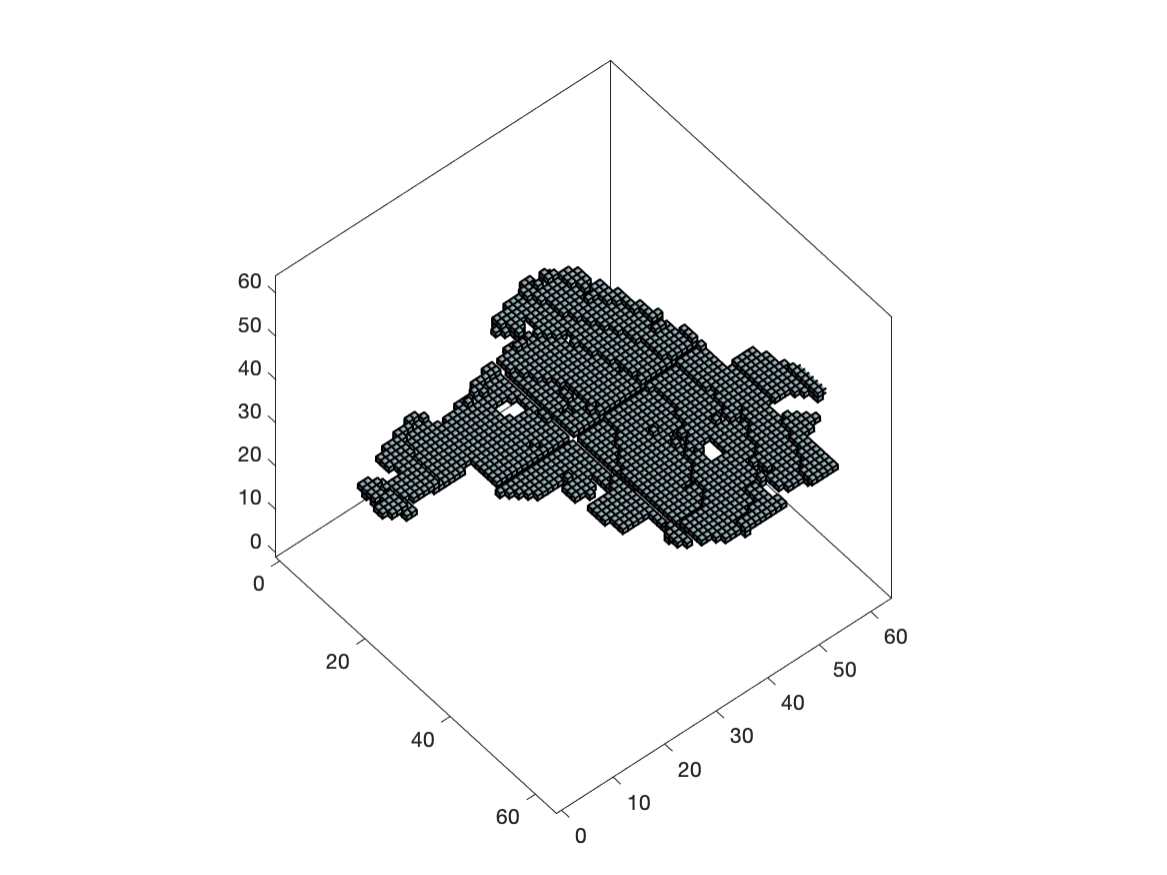} &
		\includegraphics[width=0.3\columnwidth]{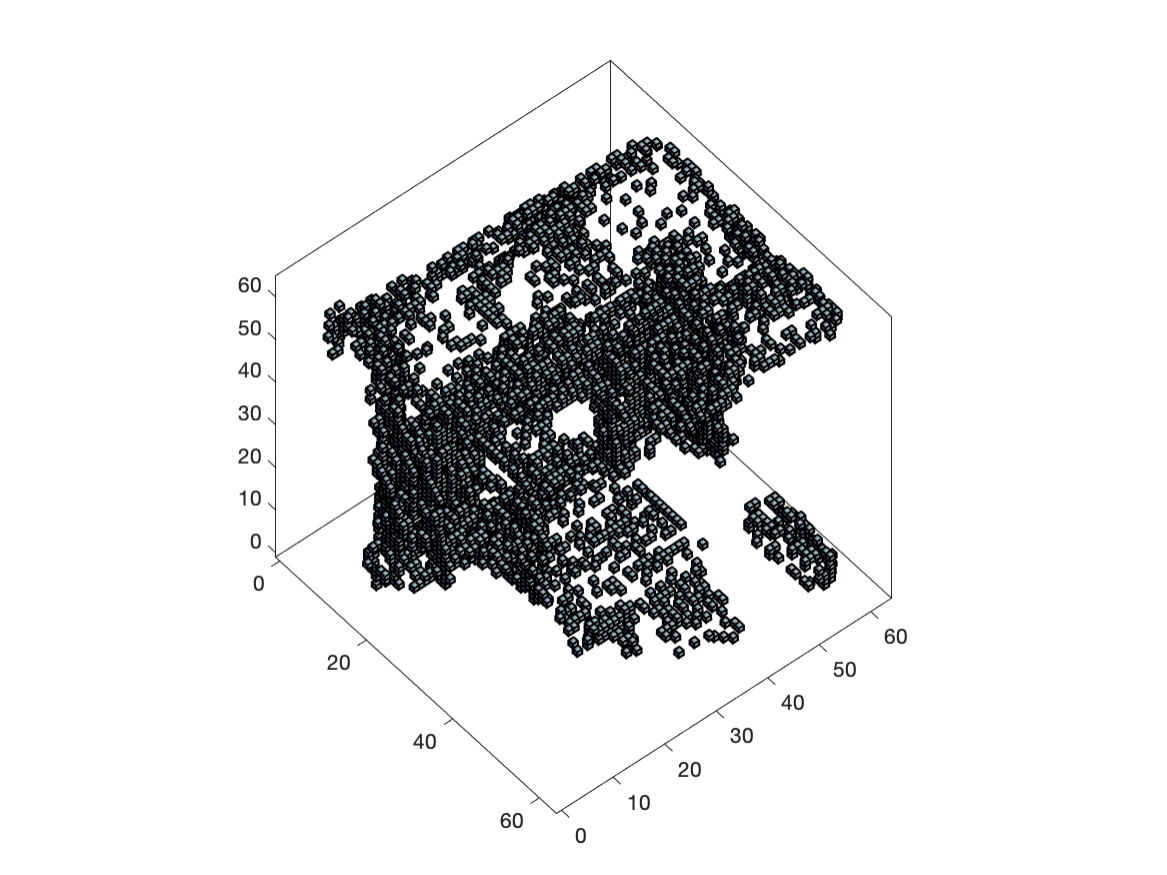}&
		\includegraphics[width=0.3\columnwidth]{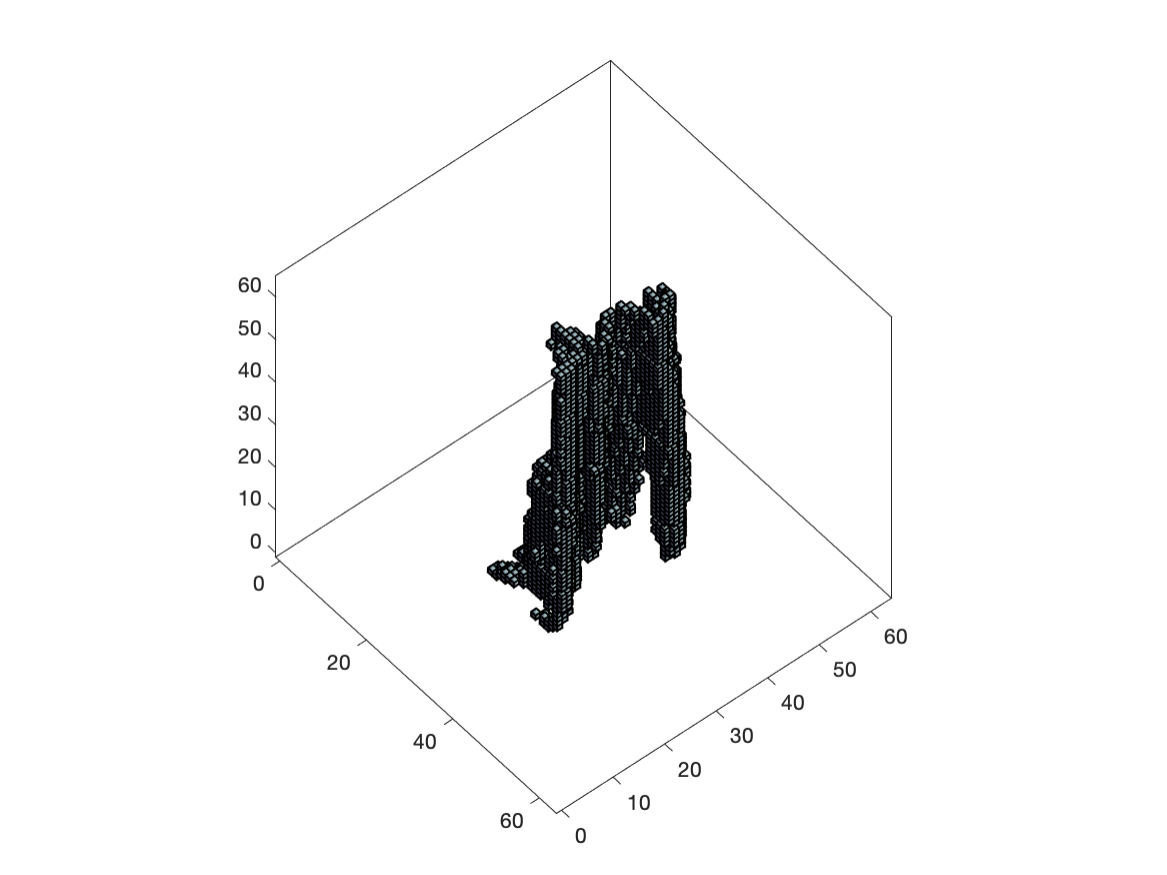}\\
		chair & bed & bathtub & floor & desk & curtain\\
		
		\includegraphics[width=0.3\columnwidth]{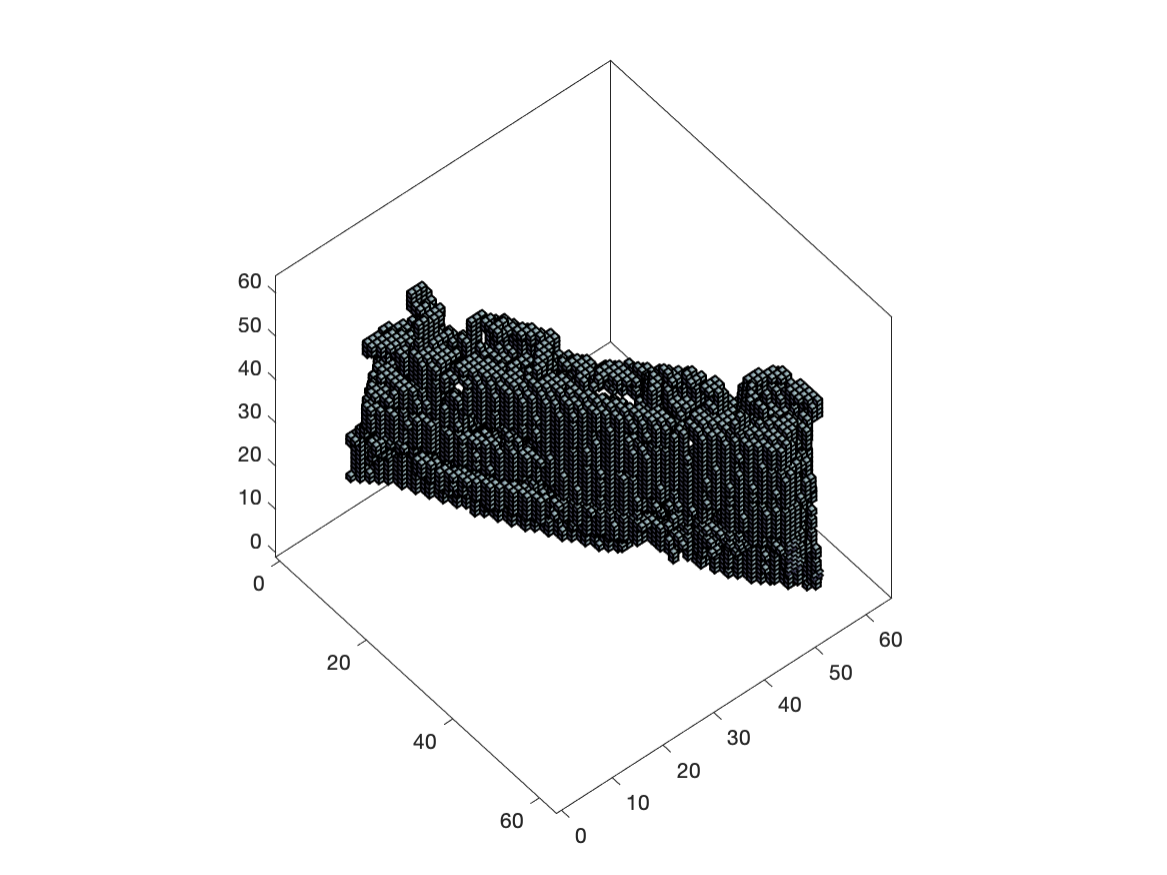} &
		\includegraphics[width=0.3\columnwidth]{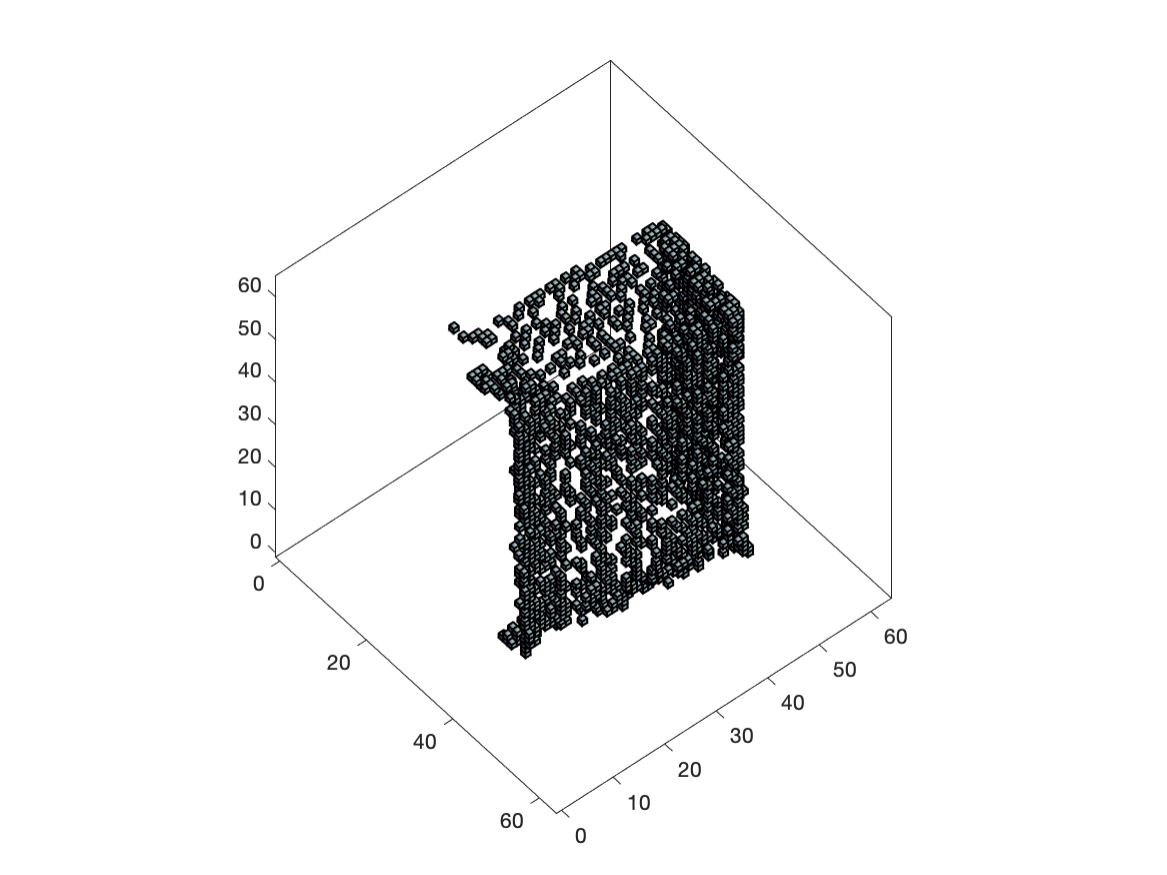}&
		\includegraphics[width=0.3\columnwidth]{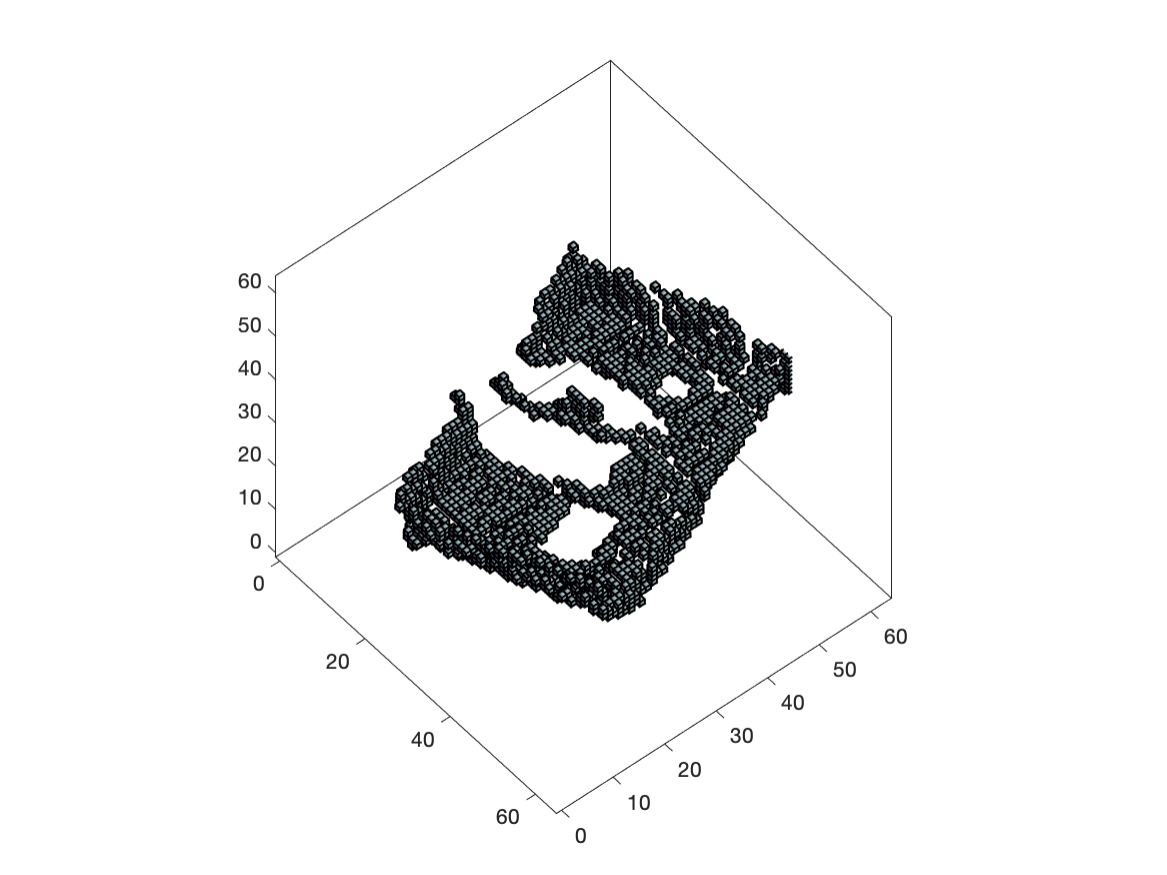}&
		\includegraphics[width=0.3\columnwidth]{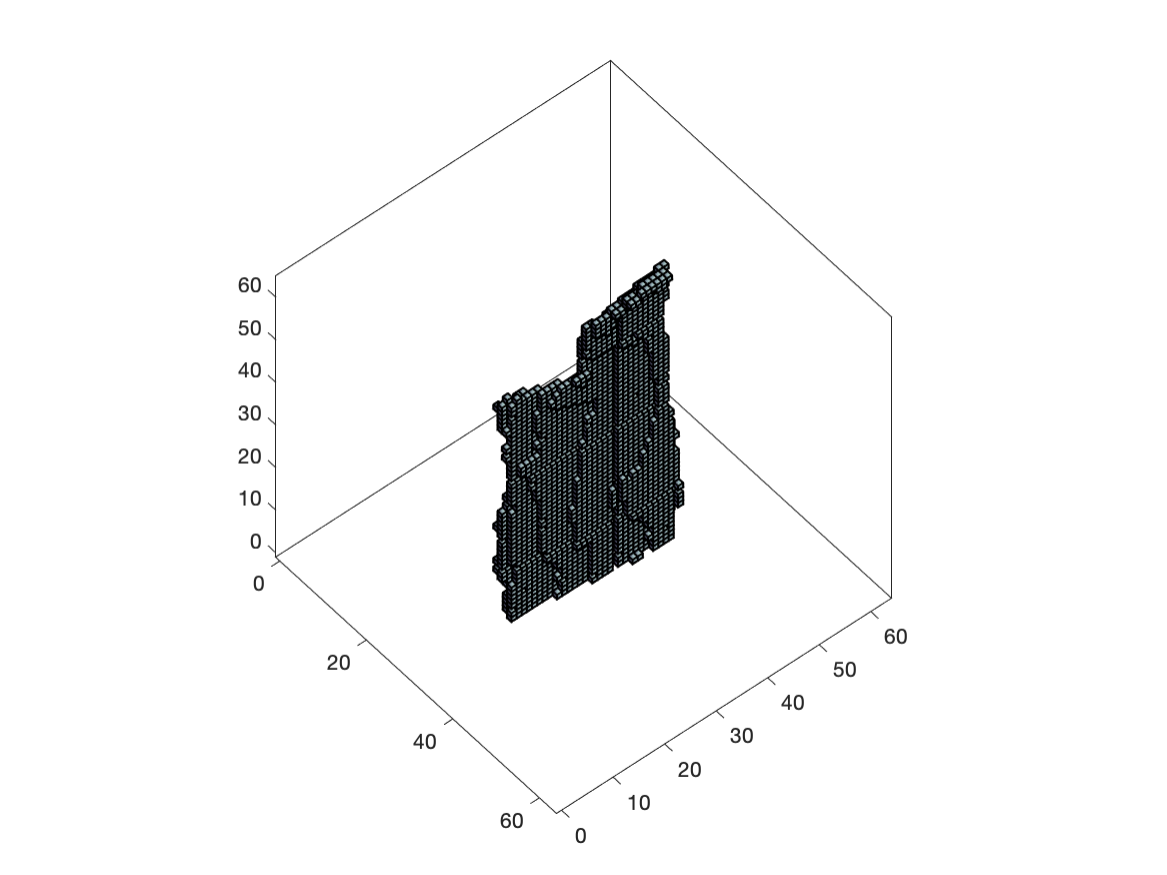} &
		\includegraphics[width=0.3\columnwidth]{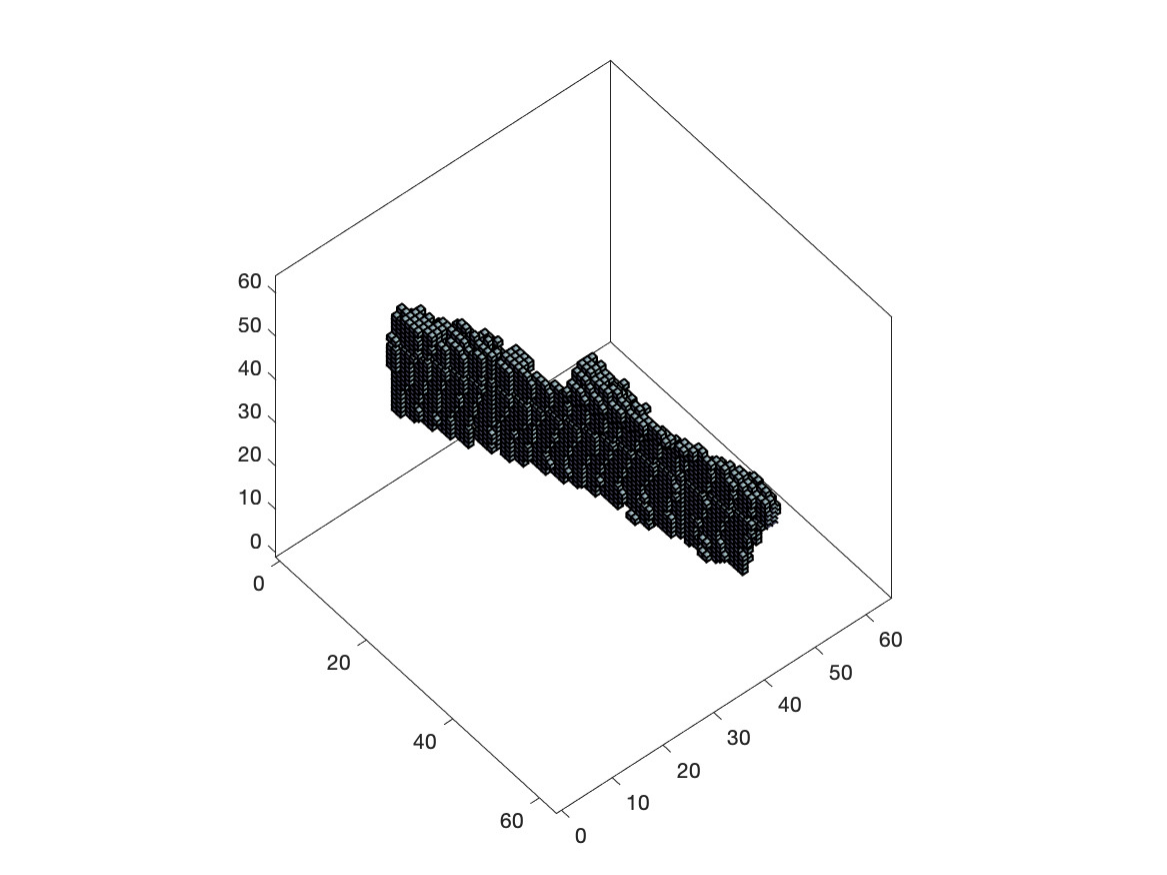}&
		\includegraphics[width=0.3\columnwidth]{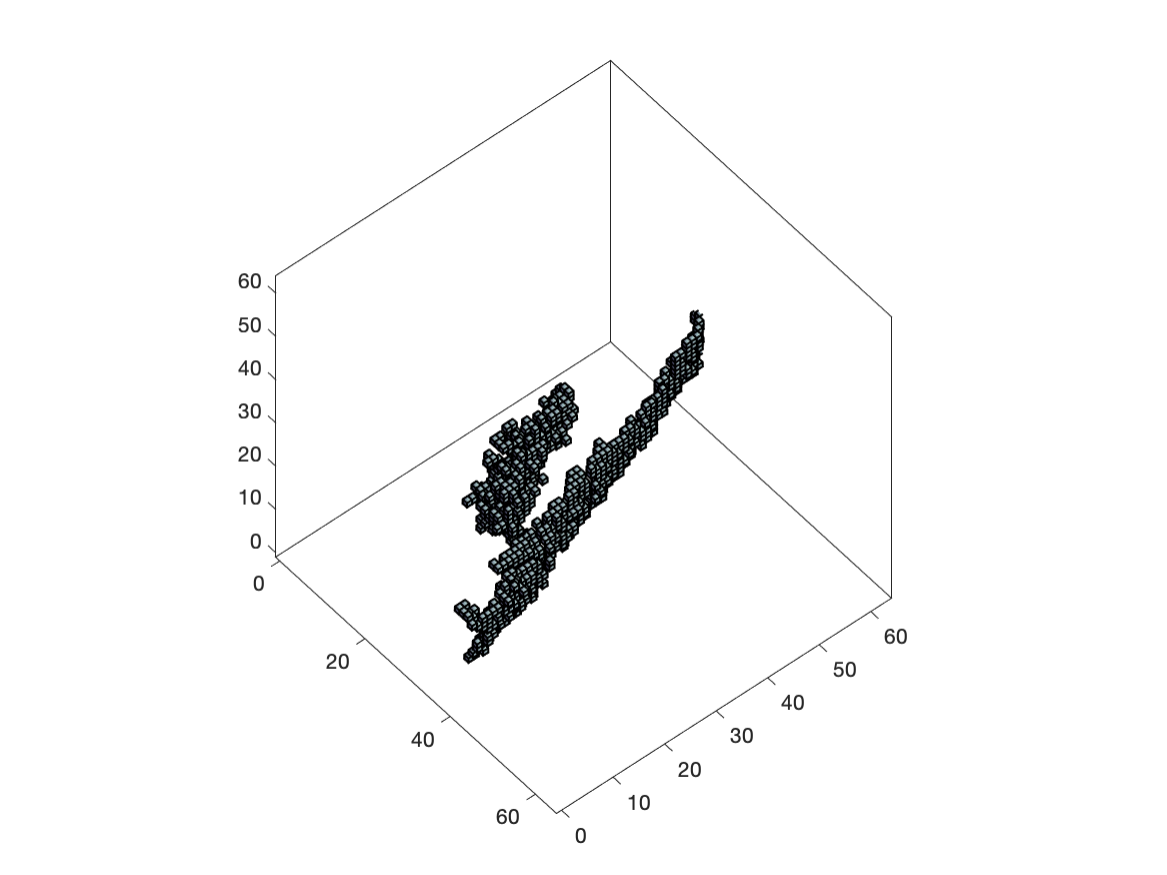}\\
		bookshelf & cabinet & counter & door & wall & picture\\
		
		\includegraphics[width=0.3\columnwidth]{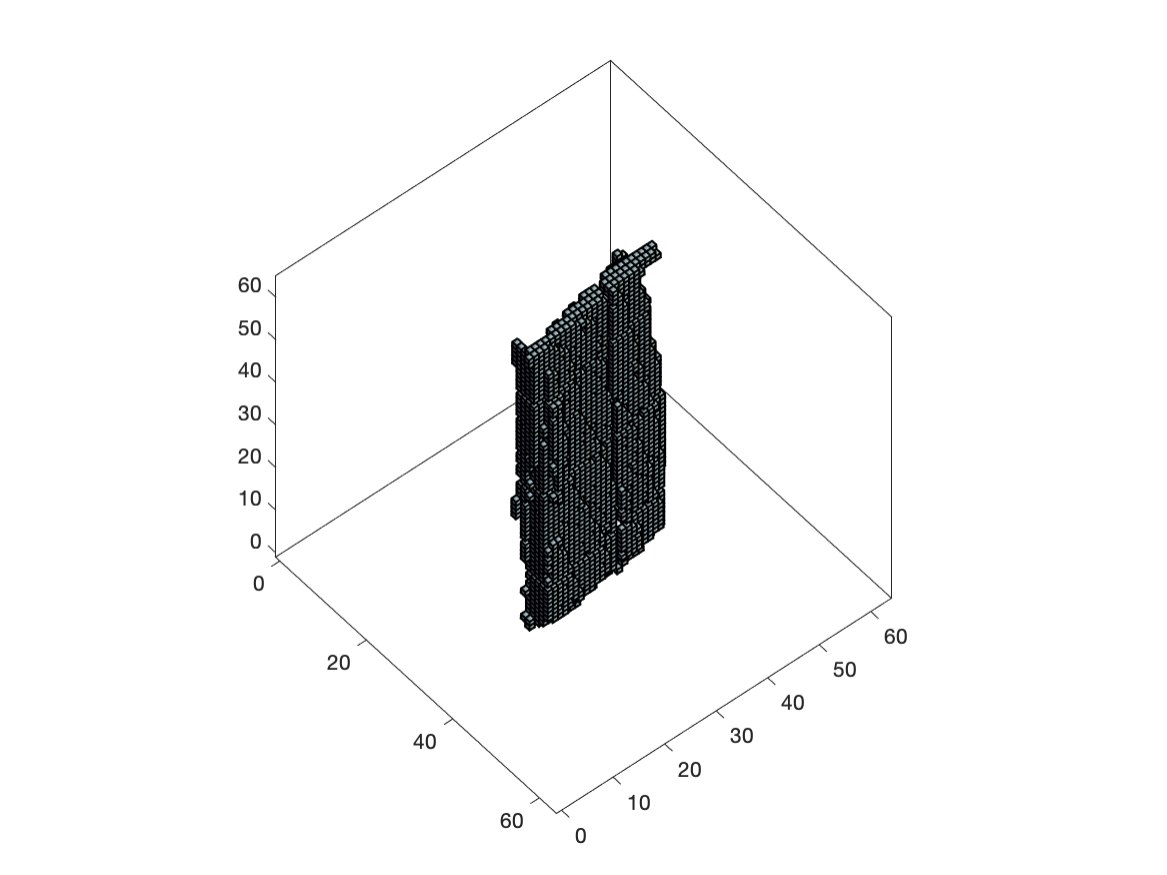} &
		\includegraphics[width=0.3\columnwidth]{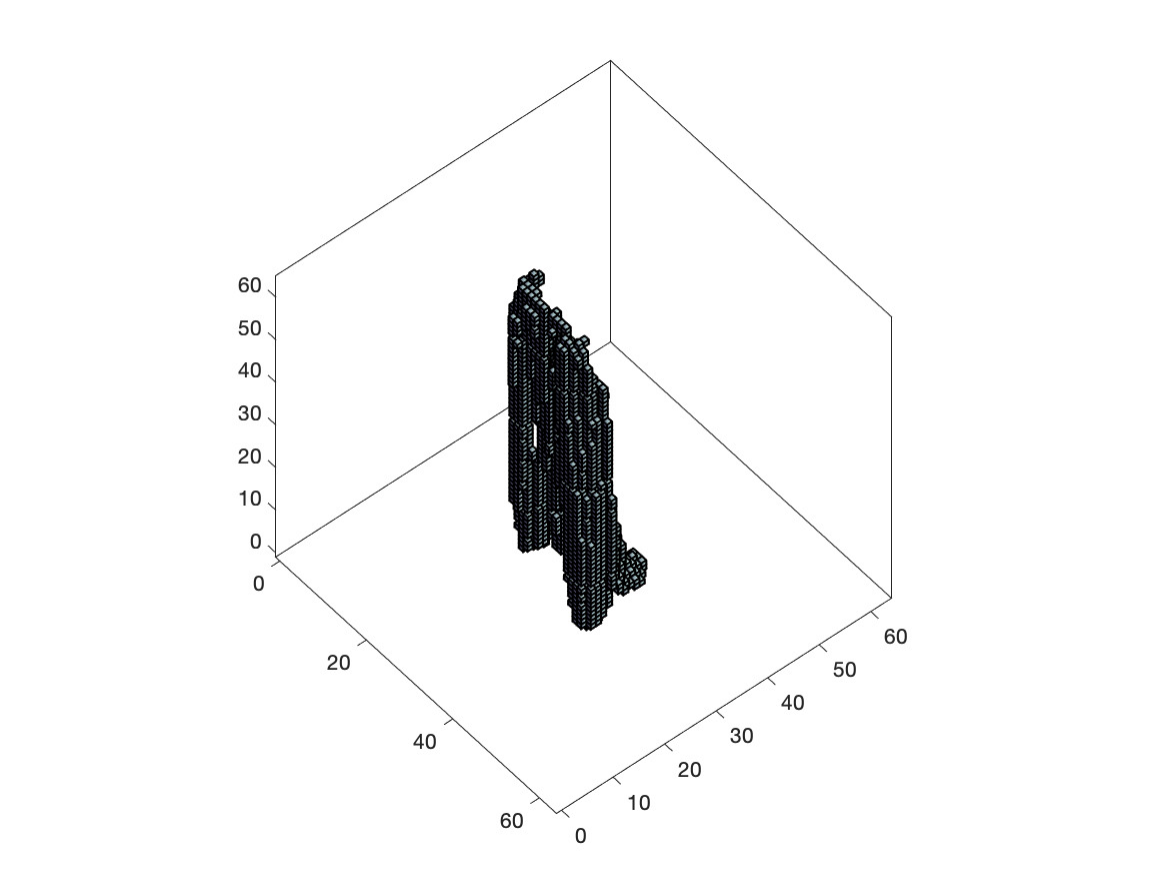}&
		\includegraphics[width=0.3\columnwidth]{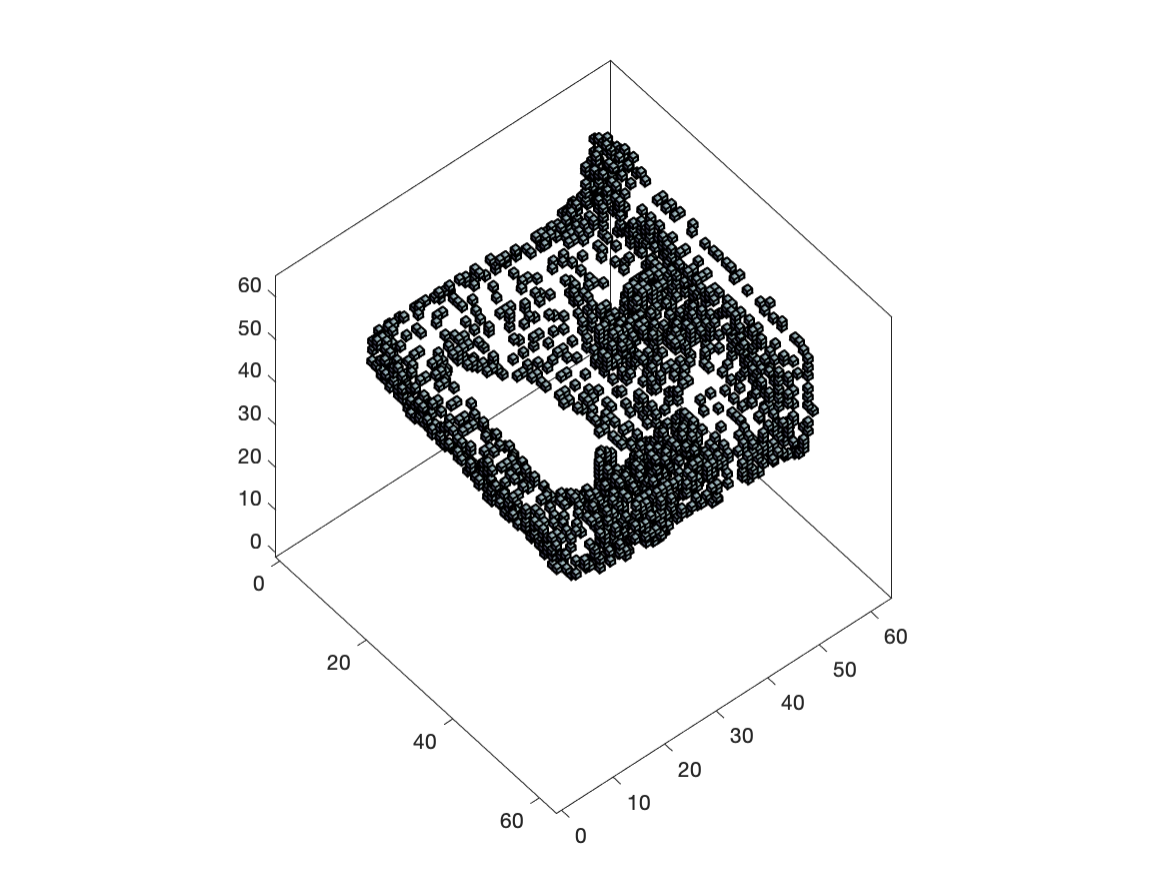}&
		\includegraphics[width=0.3\columnwidth]{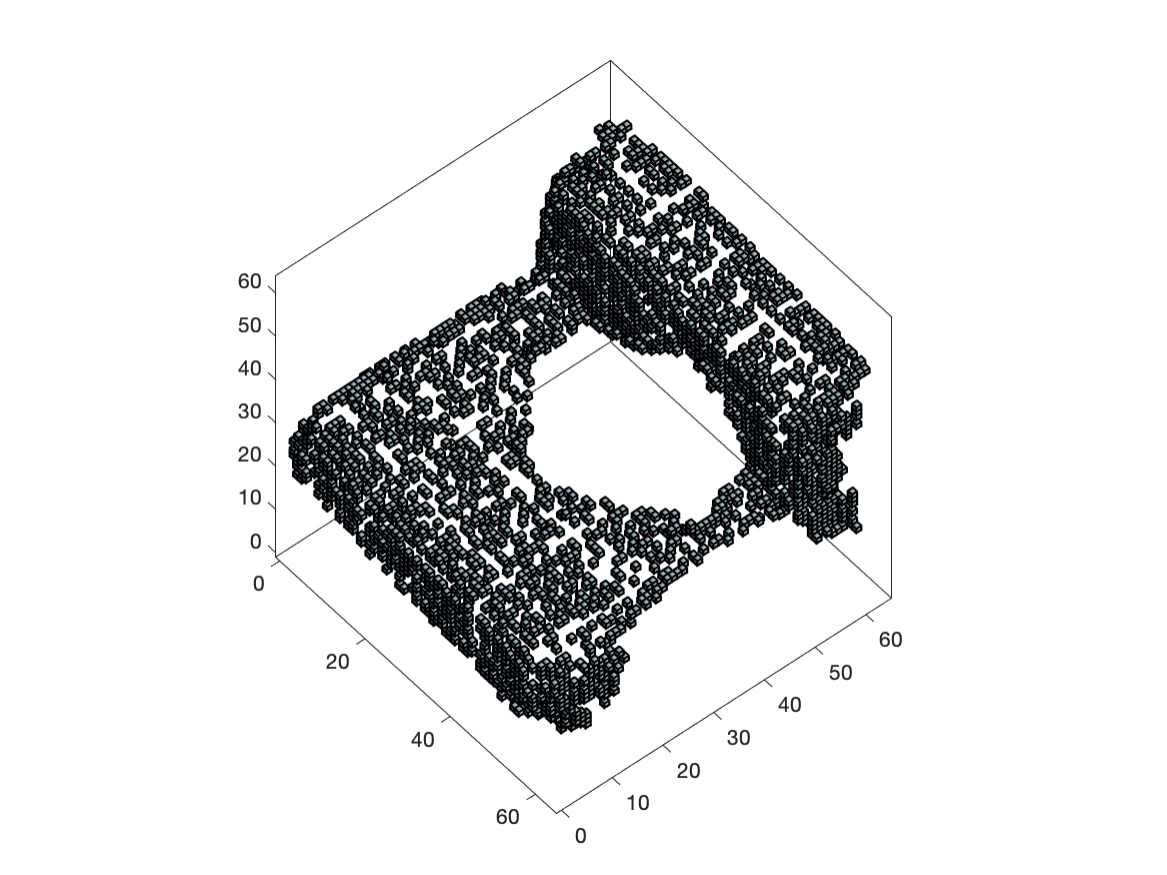} &
		\includegraphics[width=0.3\columnwidth]{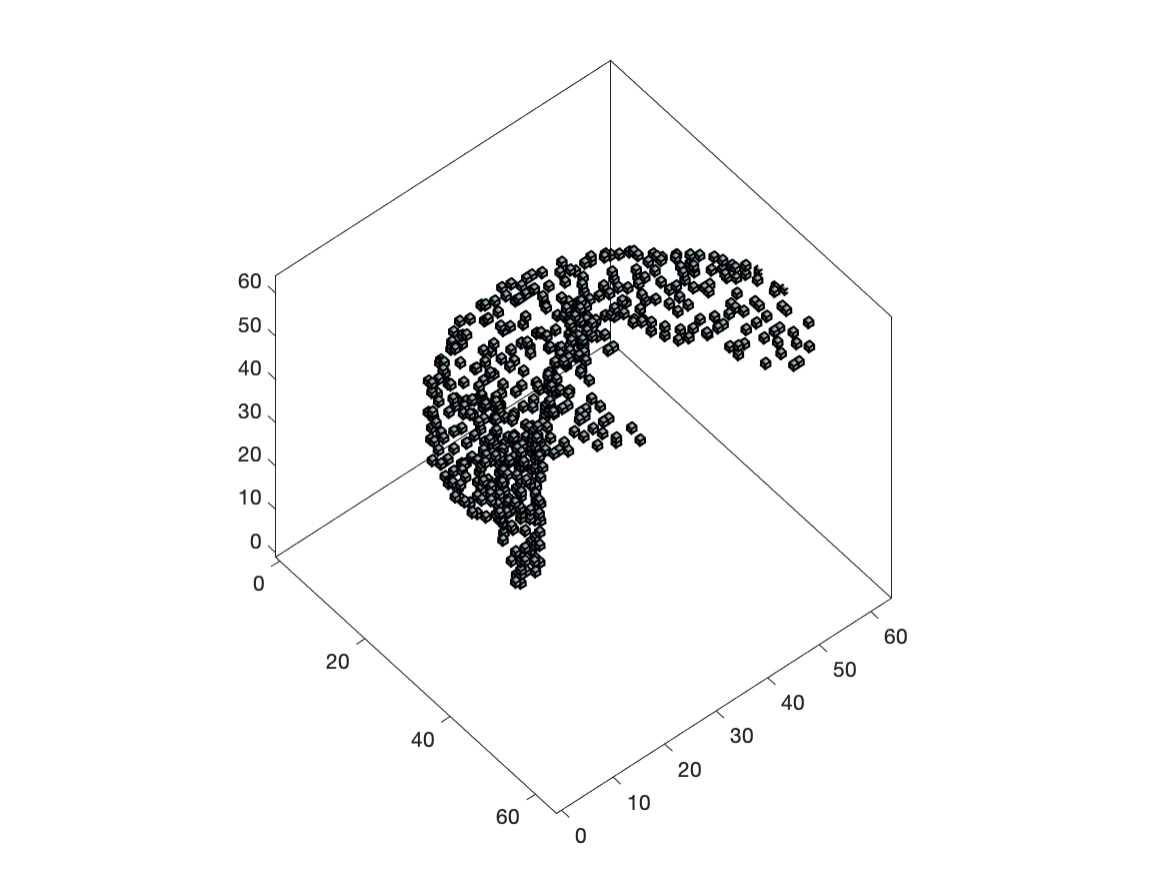}&
		\includegraphics[width=0.3\columnwidth]{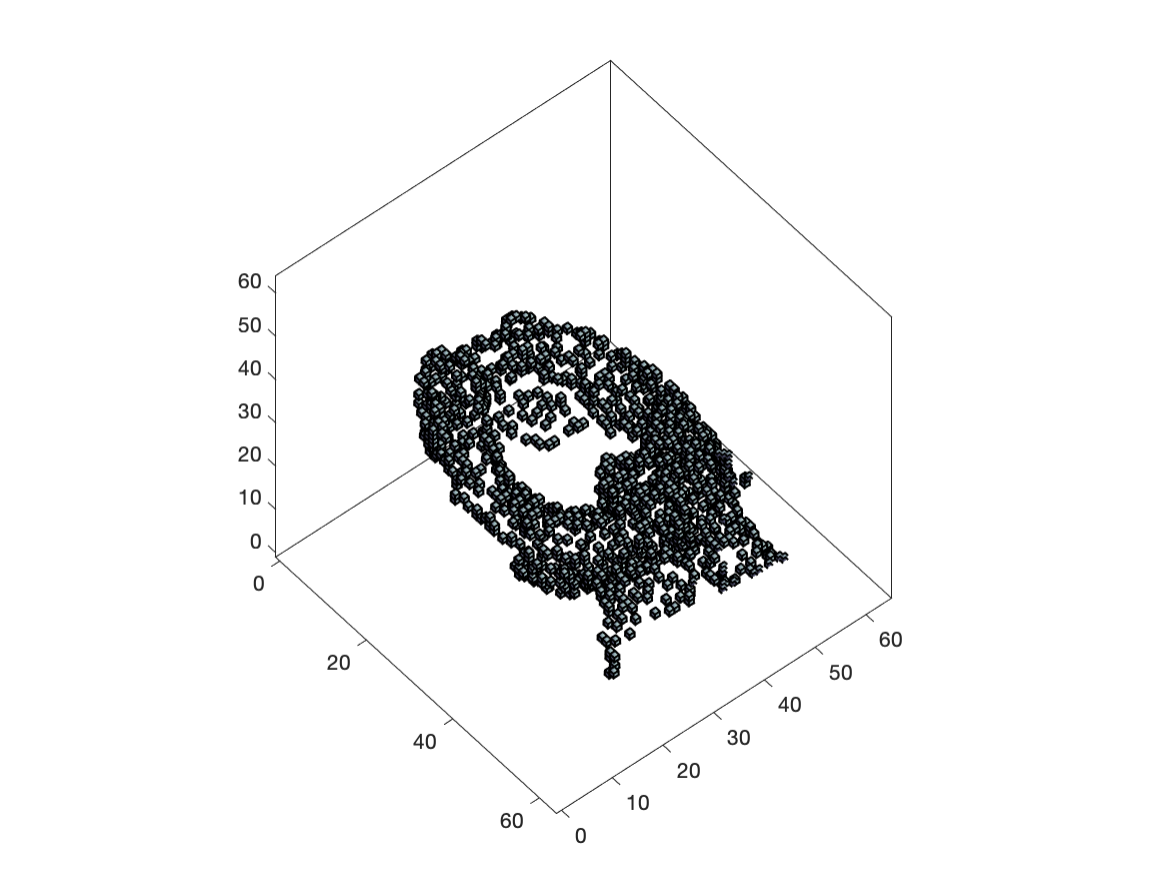}\\
		refrigerator & shower curtain & sink & sofa & table & toilet\\
	\end{tabular}
	\caption{Examples from ScanNet, at voxel resolution
		$64^3$. The sparsity is clearly visible for all objects, no
		matter what their size and shape.}
	\label{fig:scannet_data}
\end{figure*}

\subsection{Classification performance on ScanNet}

\begin{figure}[t!]
	\centering
	
	\includegraphics[height=0.3\textwidth,keepaspectratio]{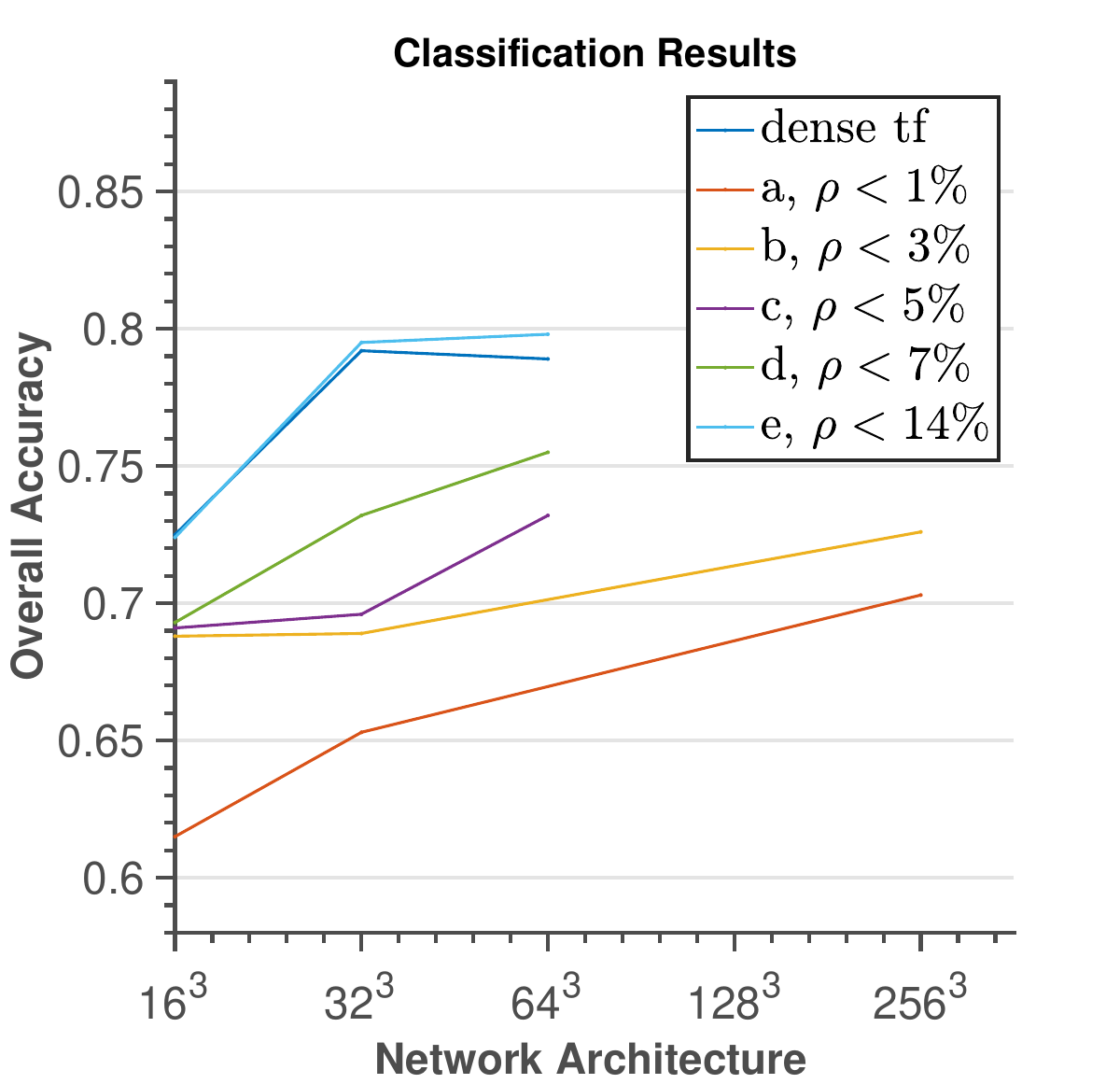}
	
	\caption{Test accuracy for 3d instance classification on
		Scannet. $\rho$ defines the upper density bound of the first
		layer, see appendix for detailed specifications. Extreme
		sparsity does decrease performance a bit, but still gives
		very reasonable results. Enforcing a moderate amount of
		sparsity does not seem to hurt the classification. At high
		resolutions, only sparse networks are tractable.}
	\label{fig:scannet2}
\end{figure}

Additionally, we also test our sparse 3d network on the Scannet
dataset. ScanNet provides 3D scans of indoor environments, acquired
with a moving RGB-D sensor.
In correspondence with the Modelnet experiments, we construct an
object classification task, by extracting point clouds of all
furniture objects in the dataset, using the ground truth per-point and
instance annotations.%
\footnote{It is not obvious how to conduct per-point classification of
	complete scans, since this would require sparse upsampling to get
	back to the input resolution.} %
The ScanNet evaluation uses 19 different classes of furniture (like
bathtub, bed, chair, desk,...), plus a rejection class ``other
furniture'', which we discard for our experiments. Our test thus is a
19-way classification task, where the input are voxel occupancies
derived from point clouds that may be incomplete and noisy, but
contain only points from the target class.
70\% of the dataset are randomly picked as training data, the rest is
used for validation. The train/test split remains the same for all
evaluated networks.
Results are given in Figures~\ref{fig:modelnet40} and \ref{fig:scannet2}.
We run experiments with: $r\in\{16, 32, 64, 256\}$. On the highest
resolution we did not test the dense CNN, since it is already close to
intractable: batch size has to be reduced to 2 even for sparse version, still a single 
epoch
runs for $>11$ hours, which means $>6$ weeks to convergence. The
sparse network at resolution $256$ trains more than $10\times$ faster.
The performance of our network is almost the same as
that of both the equivalent dense network and Octnet (in fact it is
slightly better).

\section{Conclusion}
\label{sec:conclusion}

We have proposed novel neural network mechanisms that exploit and
encourage sparsity in both feature maps and model parameters.
When the input data exhibits a substantial degree of sparsity, our
novel sparse layers and back-propagation rule significantly reduce
\emph{(i)} memory footprint and \emph{(ii)} runtime of convolutional
layers at practically useful resolutions.
Moreover, our approach guarantees upper bounds on the memory
requirements and runtime of the network.
For classification tasks the performance of our sparse network is
comparable to its dense counterpart, as well as to Octnet. In future
work it will be interesting to employ sparsity also for other tasks.
Our implementation is fully compatible with Tensorflow and has been
released as open-source code.
We hope that hardware support for sparse convolutions will improve
further on future consumer GPUs, as demonstrated by
\cite{parashar2017scnn}; thus further boosting the performance of
sparse, high-dimensional CNNs.

\bibliographystyle{apalike}
\bibliography{references}

\newpage

\appendix

\section*{Appendix: Network architectures}

Table \ref{tab:network_architectures} shows the network architectures
of our experiments. Depending on the type of data and the goal of the
experiment, we used the following network specifications (both for the
dense and the sparse version, where applicable):
\begin{itemize}
	\item
	For MNIST, we run OctNet3-$24^2$ with $10$ output classes. $\rho_{21}$
	is set as specified in the paper text and $\rho_{22} = 2 \cdot
	\rho_{21}$.\vspace{0.5em}
	\item
	For Modelnet40 we have $c=40$ different output classes and employ the
	following variants:\vspace{0.5em}
	\begin{enumerate}
		\item
		For the regularisation experiment (Figure~\ref{fig:pruning}) we use
		OctNet3-$64^3$ with $\rho_{41} = 0.06$, $\rho_{42} = 0.14$, $\rho_{43}
		= 0.33$.\vspace{0.5em}
		\item
		Figure~\ref{fig:modelnet40} \emph{(left)} has been generated with
		OctNet3-$16^3$ with varying upper bounds $\rho_{11} = \{0.06, 0.12,
		0.33, 1 \}$, $\rho_{12} = \{0.12, 0.24, 0.33, 1\}$.\vspace{0.5em}
		\item
		For Figure~\ref{fig:modelnet40} \emph{(right)} the following networks
		were used: OctNet3-$16^3$ with $\rho_{11} = \rho_{12} = 1$;
		OctNet3-$32^3$ with $\rho_{31} = 0.14$, $\rho_{32} = 0.33$,
		$\rho_{33} = 0.66$; OctNet3-$64^3$ with $\rho_{41} = 0.06$,
		$\rho_{42} = 0.14$, $\rho_{43} = 0.33$; OctNet3-$128^3$ with
		$\rho_{51} = 0.02$, $\rho_{52} = 0.06$, $\rho_{53} = 0.14$.
	\end{enumerate}
	\vspace{0.5em}
	\item For ScanNet, we use the same architectures as for
	Modelnet, but setting the number of output classes to $c=19$.
	\begin{enumerate}
		\item
		Figure~\ref{fig:modelnet40} was generated with OctNet3-$16^3$ with
		$\rho_{11} = 0.14$, $\rho_{12} = 0.33$; OctNet3-$32^3$ with
		$\rho_{31} = 0.14$, $\rho_{32} = 0.33$, $\rho_{33} = 0.66$;
		OctNet3-$64^3$ with $\rho_{41} = 0.06$, $\rho_{42} = 0.14$,
		$\rho_{43} = 0.33$.
		\item Figure~\ref{fig:scannet2} was generated with:
		\begin{enumerate}
			\item OctNet3-$16^3$ with
			$\rho_{11} = 0.01$, $\rho_{12} = 0.03$; OctNet3-$32^3$ with
			$\rho_{31} = 0.01$, $\rho_{32} = 0.03$, $\rho_{33} = 0.03$;
			OctNet3-$256^3$ with $\rho_{61} = 0.01$, $\rho_{62} = 0.03$,
			$\rho_{63} = 0.07$.
			\item OctNet3-$16^3$ with
			$\rho_{11} = 0.03$, $\rho_{12} = 0.05$; OctNet3-$32^3$ with
			$\rho_{31} = 0.03$, $\rho_{32} = 0.05$, $\rho_{33} = 0.05$;
			OctNet3-$256^3$ with $\rho_{61} = 0.03$, $\rho_{62} = 0.05$,
			$\rho_{63} = 0.07$.
			\item OctNet3-$16^3$ with
			$\rho_{11} = 0.05$, $\rho_{12} = 0.07$; OctNet3-$32^3$ with
			$\rho_{31} = 0.05$, $\rho_{32} = 0.07$, $\rho_{33} = 0.07$;
			OctNet3-$64^3$ with
			$\rho_{41} = 0.05$, $\rho_{42} = 0.07$, $\rho_{43} = 0.14$.
			\item OctNet3-$16^3$ with
			$\rho_{11} = 0.07$, $\rho_{12} = 0.09$; OctNet3-$32^3$ with
			$\rho_{31} = 0.07$, $\rho_{32} = 0.09$, $\rho_{33} = 0.09$; 
			OctNet3-$64^3$ with
			$\rho_{41} = 0.07$, $\rho_{42} = 0.09$, $\rho_{43} = 0.18$.
			\item OctNet3-$16^3$ with
			$\rho_{11} = 0.14$, $\rho_{12} = 0.33$; OctNet3-$32^3$ with
			$\rho_{31} = 0.14$, $\rho_{32} = 0.33$, $\rho_{33} = 0.66$; 
			OctNet3-$64^3$ with
			$\rho_{41} = 0.6$, $\rho_{42} = 0.14$, $\rho_{43} = 0.33$.
		\end{enumerate}
	\end{enumerate}
	
\end{itemize}

\begin{table*}[htb]
	\centering
	\footnotesize
	\begin{tabular}{|c|c|c|c|c|c|}
		\hline
		OctNet3-$16^3$ & OctNet3-$24^2$ & OctNet3-$32^3$ & OctNet3-$64^3$ & 
		OctNet3-$128^3$ & OctNet3-$256^3$\\
		\hline
		conv($1$, $8$, $\rho_{11}$) & conv($1$, $8$, $\rho_{21}$) & conv($1$, 
		$8$, $\rho_{31}$) & conv($1$, $8$, $\rho_{41}$) & conv($1$, $8$, 
		$\rho_{51}$) & conv($1$, $8$, $\rho_{61}$) \\
		conv($8$, $8$, $\rho_{11}$) & conv($8$, $8$, $\rho_{21}$) & conv($8$, 
		$8$, $\rho_{31}$) & conv($8$, $8$, $\rho_{41}$) & conv($8$, $8$, 
		$\rho_{51}$) & conv($8$, $8$, $\rho_{61}$) \\
		conv($8$, $8$, $\rho_{11}$) & conv($8$, $8$, $\rho_{21}$) & conv($8$, 
		$8$, $\rho_{31}$) & conv($8$, $8$, $\rho_{41}$) & conv($8$, $8$, 
		$\rho_{51}$) & conv($8$, $8$, $\rho_{61}$) \\
		maxPooling(2) & maxPooling(2) & maxPooling(2) & maxPooling(2) & 
		maxPooling(2) & maxPooling(2) \\
		\hline
		conv($8$, $16$, $\rho_{12}$) & conv($8$, $16$, $\rho_{22}$) & 
		conv($8$, $16$, $\rho_{32}$) & conv($8$, $16$, $\rho_{42}$) & 
		conv($8$, $16$, $\rho_{52}$) &  conv($8$, $16$, $\rho_{62}$)\\
		conv($16$, $16$, $\rho_{12}$) & conv($16$, $16$, $\rho_{22}$) & 
		conv($16$, $16$, $\rho_{32}$) & conv($16$, $16$, $\rho_{42}$) & 
		conv($16$, $16$, $\rho_{52}$) & conv($16$, $16$, $\rho_{62}$) \\
		conv($16$, $16$, $\rho_{12}$) & conv($16$, $16$, $\rho_{22}$) & 
		conv($16$, $16$, $\rho_{32}$) & conv($16$, $16$, $\rho_{42}$) & 
		conv($16$, $16$, $\rho_{52}$) & conv($16$, $16$, $\rho_{62}$) \\
		sparseToDense() & sparseToDense() & maxPooling(2) & maxPooling(2) & 
		maxPooling(2) & maxPooling(2) \\
		\hline
		& & conv($16$, $24$, $\rho_{33}$) & conv($16$, $24$, $\rho_{43}$) & 
		conv($16$, $24$, $\rho_{53}$) &  conv($16$, $24$, $\rho_{63}$)\\
		& & conv($24$, $24$, $\rho_{33}$) & conv($24$, $24$, $\rho_{43}$) & 
		conv($24$, $24$, $\rho_{53}$) & conv($24$, $24$, $\rho_{63}$) \\
		& & conv($24$, $24$, $\rho_{33}$) & conv($24$, $24$, $\rho_{43}$) & 
		conv($24$, $24$, $\rho_{53}$) & conv($24$, $24$, $\rho_{63}$) \\
		& &  & sparseToDense() & sparseToDense() & sparseToDense() \\
		\hline
		& & & maxPooling(2) & maxPooling(2) & maxPooling(2) \\
		& & & conv($24$, $32$) & conv($24$, $32$) & conv($24$, $32$) \\
		& & & conv($32$, $32$) & conv($32$, $32$) & conv($32$, $32$) \\
		& & & conv($32$, $32$) & conv($32$, $32$) & conv($32$, $32$) \\
		\hline
		& & & & maxPooling(2) & maxPooling(2) \\
		& & & & conv($32$, $40$) & conv($32$, $40$) \\
		& & & & conv($40$, $40$) & conv($40$, $40$) \\
		& & & & conv($40$, $40$) & conv($40$, $40$) \\
		\hline
		& & & & & maxPooling(2) \\
		& & & & & conv($40$, $48$)\\
		& & & & & conv($48$, $48$) \\
		& & & & & conv($48$, $48$) \\
		\hline
		\multicolumn{6}{|c|}{dropout($0.5$)} \\
		\multicolumn{6}{|c|}{fully-connected($1024$)} \\
		\multicolumn{6}{|c|}{fully-connected($c$)} \\
		
		\hline
	\end{tabular}
	\caption{In our evaluation, we use different Octnet3 network
		architectures, similar to those also used by Riegler et 
		al.~\cite{Riegler2017OctNet}}
	\label{tab:network_architectures}
\end{table*}

\end{document}